\documentclass[times,twocolumn,final]{elsarticle}

\usepackage{medima}
\usepackage{framed,multirow}

\usepackage{amsmath,amssymb,amsfonts}
\usepackage{latexsym}

\usepackage{url}
\usepackage{xcolor}
\usepackage[colorlinks, citecolor=blue]{hyperref}
\usepackage{algorithmic}
\usepackage{graphicx}
\usepackage{textcomp}
\usepackage{bm}
\usepackage{makecell}
\usepackage{multirow}
\usepackage{bbding}
\usepackage{subfig}
\usepackage[switch]{lineno}
\usepackage{booktabs}
\usepackage{threeparttable}
\usepackage{array}

\definecolor{newcolor}{rgb}{.8,.349,.1}
\journal{Medical Image Analysis}

\begin{document}

\verso{Lin \textit{et~al.}}

\begin{frontmatter}

\title{Instance-Aware Robust Consistency Regularization for Semi-Supervised Nuclei Instance Segmentation}

\author[1]{Zenan Lin\fnref{equal}}
\author[2]{Wei Li\fnref{equal}}
\author[3]{Jintao Chen}
\author[1]{Zihao Wu}
\author[1,6,7]{Wenxiong Kang}
\author[4]{Changxin Gao}
\author[5]{Liansheng Wang}
\author[1,6]{Jin-Gang Yu\corref{cor1}}

\cortext[cor1]{Corresponding authors: Jin-Gang Yu (jingangyu@scut.edu.cn)}
\fntext[equal]{The first two authors contributed equally to this work.}

\address[1]{School of Automation Science and Engineering, South China University of Technology, Guangzhou 510641, China}
\address[2]{Department of Breast and Thyroid Surgery, the Second Affiliated Hospital, University of South China, Hengyang 421001, China}
\address[3]{Fangxin Cooperation, Guangzhou 510705, China}
\address[4]{School of Artificial Intelligence and Automation, Huazhong University of Science and Technology, Wuhan 430074, China}
\address[5]{School of Informatics, Xiamen University, Xiamen 361005, China}
\address[6]{Pazhou  Laboratory, Guangzhou 510335, China}
\address[7]{School of Future Technology, South China University of Technology, Guangzhou 510641, China}


\begin{abstract}
Nuclei instance segmentation in pathological images is crucial for downstream tasks such as tumor microenvironment analysis. However, the high cost and scarcity of annotated data limit the applicability of fully supervised methods, while existing semi-supervised methods fail to adequately regularize consistency at the instance level, lack leverage of the inherent prior knowledge of pathological structures, and are prone to introducing noisy pseudo-labels during training. In this paper, we propose an Instance-Aware Robust Consistency Regularization Network (IRCR-Net) for accurate instance-level nuclei segmentation. Specifically, we introduce the Matching-Driven Instance-Aware Consistency (MIAC) and Prior-Driven Instance-Aware Consistency (PIAC) mechanisms to refine the nuclei instance segmentation result of the teacher and student subnetwork, particularly for densely distributed and overlapping nuclei. We incorporate morphological prior knowledge of nuclei in pathological images and utilize these priors to assess the quality of pseudo-labels generated from unlabeled data. Low-quality pseudo-labels are discarded, while high-quality predictions are enhanced to reduce pseudo-label noise and benefit the network's robust training.  Experimental results demonstrate that the proposed method significantly enhances semi-supervised nuclei instance segmentation performance across multiple public datasets compared to existing approaches, even surpassing fully supervised methods in some scenarios.
\end{abstract}


\end{frontmatter}


\section{Introduction}

Nuclei instance segmentation enables the quantization of cellular morphological features from pathological images, which thus plays a critical role in many computational pathology tasks, such as tumor microenvironment analysis, immune scoring, and prognosis prediction \citep{litjens2017survey, pan2023smile, lou2024structure, graham2019mild, di2022generating}. Notice that, by nuclei instance segmentation, we mean the task of localizing the spatial extents of nucleus individuals only, without further classifying the nuclei. 

One major challenge with accurate nuclei instance segmentation lies in the lack of manual annotation \citep{graham2019hover, bilodeau2022microscopy}. According to the previous studies \citep{englbrecht2021automatic, graham2021lizard, lou2022pixel}, annotating a single nucleus takes an average of 8.43 seconds, and a single whole-slide image (WSI) typically contains hundreds of thousands of nuclei; it is hence extremely expensive to acquire sufficient annotated data for model training. To conquer this challenge, researchers have developed semi-supervised learning (SSL) methods, which utilize both limited labeled data and abundant unlabeled data to boost model performance \citep{tarvainen2017mean, yu2019uncertainty, mittal2019semi, xie2021intra}. In the present work, we concentrate on a particular category of SSL methods, i.e., consistency regularization methods \citep{chen2021semi}. Most commonly, these methods follow a Teacher-Student framework \citep{tarvainen2017mean}, where the teacher model and the student model respectively process a permutation of the unlabeled sample, and a consistency regularization term is imposed between the two predictions. Central to these methods is how to establish the consistency regularization term.

\begin{figure}[]
	\centering
	\includegraphics[width=1.0\linewidth]{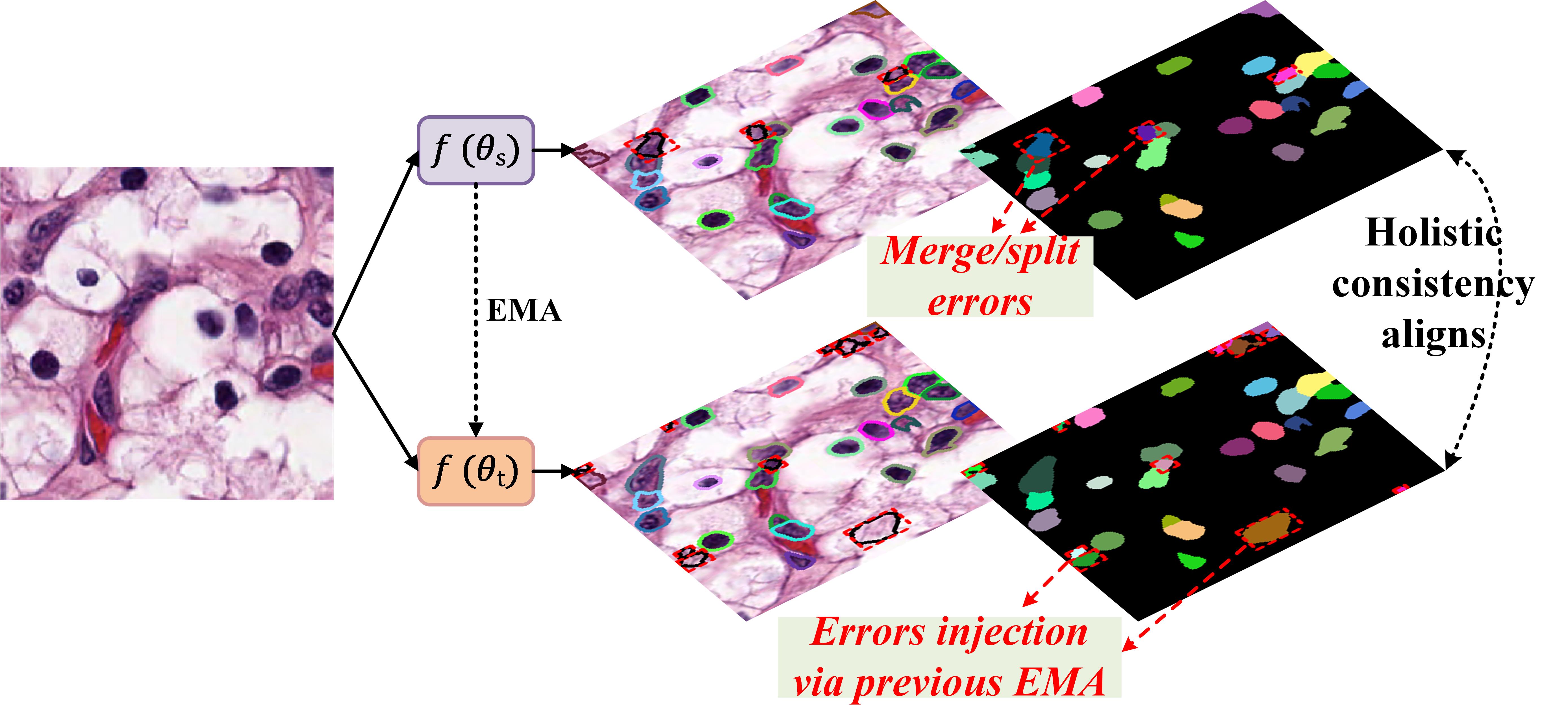}
	\caption{Holistic consistency aligns global maps and thus admits merge/split errors and error injection via previous EMA.}
	\label{fig:global_vs_instance}
\end{figure}

Consistency regularization SSL methods have not yet been fully exploited in the context of nuclei instance segmentation. Among the few previous works, Wu et al. \citep{wu2022cross} proposed a cross-patch dense contrastive learning. Jin et al. \citep{jin2024inter} proposed an inter- and intra-uncertainty regularization framework with a two-stage pseudo-mask guided feature aggregation network. As illustrated in Fig. \ref{fig:global_vs_instance}, these works share a common limitation, i.e., their consistency regularization terms are defined by holistically contrasting the predictions (either the intermediate feature maps or the final outputs) of the teacher model and the student model with a certain distance measure. With such holistic consistency regularization, the incorrect results generated by the student model will be undesirably brought into the loop to corrupt the teacher model, especially at the early iterations of the model training procedure.

To tackle this issue, this paper presents an \textit{Instance-Aware Robust Consistency Regularization Network (IRCR-Net)} for semi-supervised nuclei
instance segmentation. The basic idea of IRCR is to establish consistency regularization at the instance level, rather than holistically. For this purpose, we propose two mechanisms to make the consistency regularization instance-aware, called \textit{Matching-Driven Instance-Aware Consistency (MIAC)} and \textit{Prior-Driven Instance-Aware Consistency (PIAC)}, respectively. MIAC introduces a bipartite matching procedure to match the instances generated by the teacher model and by the student model, and only those matched instances are involved in establishing the consistency loss. Our intuition is that an instance predicted by one model that cannot find a match in the results predicted by the other model tends to be an incorrect prediction, which should be excluded from the consistency regularization for robustness. PIAC takes advantage of the prior knowledge of nuclei, in terms of the statistical distribution, to quantify the probability that each predicted instance is a nucleus. Instances predicted with low probability are considered to be incorrect and excluded from the consistency regularization to improve its robustness. The statistical prior is derived in a non-parametric fashion by using kernel density estimation over a set of well-defined handcrafted features. Both MIAC and PIAC are incorporated into a common Mean-Teacher structure, where Hover-Net \citep{graham2019hover} is taken as the fundamental segmentation network, to establish a framework for semi-supervised nuclei instance segmentation, called IRCR-Net. Experimental results demonstrate that the proposed method significantly enhances semi-supervised nuclei instance segmentation performance across multiple public datasets compared to existing approaches, even surpassing fully supervised methods in some scenarios.

In summary, the key contributions of this work are as follows:
\begin{itemize}
	\item We propose the concept of IRCR for the task of semi-supervised nuclei instance segmentation, which establishes consistency regularization terms at the instance level, rather than holistically to improve the robustness to incorrect predictions.
	
	\item We design two specific methods, termed as MIAC and PIAC, to implement the idea of IRCR.
	
	\item We build a unified framework called IRCR-Net based upon MIAC and PIAC and achieve superior performance across multiple public datasets.
\end{itemize}

\section{Related work}
\subsection{Semi-Supervised Methods}
Semi-supervised learning (SSL) mitigates the scarcity of annotated data by leveraging limited labeled samples alongside abundant unlabeled data \citep{lee2013pseudo, ouali2020semi, liu2022perturbed, mittal2019semi, sohn2020fixmatch, xie2021intra} and has garnered significant attention from researchers. Consistency regularization methods in SSL utilize network predictions under various perturbations to enforce consistency and exploit unlabeled data effectively \citep{chen2021semi, french2020semi}. Typically, SSL combines standard supervised loss terms (e.g., cross-entropy loss) with consistency loss terms to maintain model performance in data-scarce scenarios \citep{french2020semi}. Among these methods, the Mean-Teacher framework, which evolved from the $\Pi$-model and Temporal Ensembling, has become a cornerstone in SSL \citep{laine2022temporal, tarvainen2017mean}. By weighted averaging model weights, Mean-Teacher improves model robustness and test accuracy, achieving superior performance with fewer labeled samples compared to Temporal Ensembling. Derivative methods based on Mean-Teacher have demonstrated exceptional performance in medical image segmentation tasks with limited labeled data \citep{yu2019uncertainty, li2020transformation, jin2022semi}. For instance, Chen et al. \citep{chen2021semi} proposed the Cross Pseudo Supervision (CPS) method to enhance segmentation performance through cross-supervised pseudo-labeling. Shen et al. \citep{shen2023co} introduced the Uncertainty-Guided Collaborative Mean-Teacher (UCMT) approach to generate high-confidence pseudo-labels.

Despite their success, most consistency-based SSL methods fail to account for the challenges posed by nuclei's dense and overlapping nature in pathological images \citep{guo2023sac}. Nuclei instances often exhibit blurred boundaries and overlapping, making it difficult for global consistency constraints to capture fine-grained differences between instances. The previous methods relying solely on global feature map consistency lack precise alignment and fine-grained constraints for individual instances, rendering them less effective for instance-level segmentation tasks in pathology. To address these limitations, this study focuses on refining nuclei instance segmentation by leveraging instance-level consistency for precise segmentation. 

\subsection{Nuclei Instance Segmentation Methods}

Nuclei instance segmentation is pivotal in histopathological analysis and has garnered significant research interest. Traditional nuclei segmentation methods primarily rely on morphological operations or classical machine learning algorithms \citep{naik2008automated, veta2013automatic, jung2010unsupervised, yang2006nuclei, zhou2023cyclic}. For example, Jung et al. \citep{jung2010unsupervised} proposed an unsupervised Bayesian classification approach for separating overlapping nuclei. Yang et al. \citep{yang2006nuclei} introduced a marker-controlled watershed method based on mathematical morphology. However, these traditional methods generally exhibit poor generalization performance and often fail in complex scenarios. With the rapid development of deep learning, methods based on convolutional neural networks (CNNs) have become the dominant approach for nuclei instance segmentation. Graham et al. \citep{graham2019hover} proposed HoverNet, which combines foreground semantic segmentation with spatial structural information (e.g., distance maps) to distinguish nuclei effectively. Doan et al. \citep{doan2022sonnet} developed a self-guided ordinal regression neural network for simultaneous nucleus segmentation and classification, focusing on uncertain regions during training. To address blurred nucleus boundaries, Kumar et al. \citep{kumar2017dataset} introduced a boundary-aware deep learning method that emphasizes accurate boundary recognition, particularly for separating touching and overlapping nuclei. Pan et al. \citep{pan2023smile} proposed a novel coarse-to-fine marker-controlled watershed post-processing step to mitigate segmentation issues for large and indistinct nuclei. Qu et al. \citep{qu2019improving} designed a variance-constrained cross-entropy loss within a full resolution CNN to capture spatial relationships between pixels and achieve robust nucleus and gland segmentation. Some other studies \citep{chen2016dcan, french2020semi, kumar2017dataset} have improved segmentation accuracy by enhancing attention to instance boundaries.

Recently, the Segment Anything Model (SAM) and its variants have achieved remarkable success in general medical image segmentation tasks, such as CT and MRI, demonstrating strong generalization across modalities \citep{ma2024segment, wu2025medical, zhang2023customized}. However, when applied to histopathological images, especially for fine-grained nuclei instance segmentation, their performance remains limited due to the extreme heterogeneity, dense packing, and subtle boundary variations of nuclei \citep{chen2025segment}. More  recently, Chen et al. \citep{chen2025segment} attempted to address this issue by combining SAM with natural language for pathology images, showing promising improvements and indicating a potential direction for bridging foundation models and domain-specific nuclei segmentation.

However, the overlapping of nuclei, glands, and other tissue structures with blurred boundaries in histopathological images pose significant challenges for existing methods, which often fail to capture fine-grained instance differences or achieve precise boundary delineation \citep{pan2023smile, wu2022cross, nunes2025survey}. To address these problems, this study leverages instance-level consistency to enhance nuclei and their boundaries and integrates prior knowledge to refine the quality of predicted nuclei.

\section{Methodology}

\begin{figure*}
	\centering
	\setlength{\belowcaptionskip}{-0.2cm}
	\includegraphics[scale = 0.5]{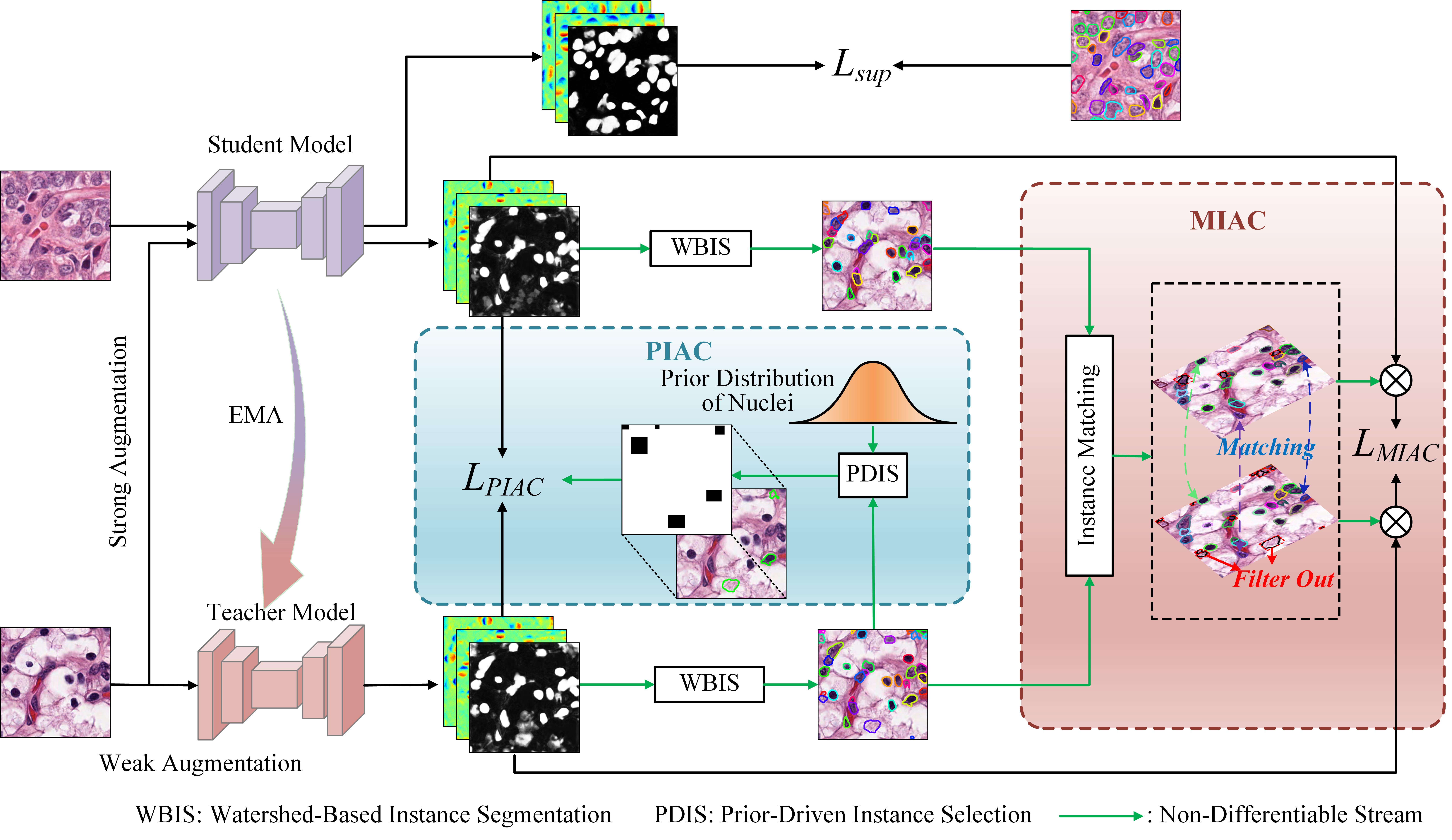}
	\caption{Overview of the proposed Instance-Aware Robust Consistency Regularization Network (IRCR-Net) for semi-supervised nuclei instance segmentation. The framework incorporates a teacher-student network architecture with two key consistency mechanisms, Matching-Driven Instance-Aware Consistency (MIAC) and Prior-Driven Instance-Aware Consistency (PIAC), to improve semi-supervised nuclei instance segmentation performance.}
	\label{fig:framework}
\end{figure*}

\subsection{Overview of the IRCR-Net Framework}

For an input histopathological image $\boldsymbol{I}$, the task of nuclei instance segmentation is to output an equally-sized instance mask image $\boldsymbol{M} \in \{0, 1, ..., N\}^{|\boldsymbol{I}|}$ labeling $N$ instances. As aforementioned, our task here is only to label the pixels belonging to each nucleus instance, without further classifying these instances. In this work, we aim to train the nuclei instance segmentation model in a semi-supervised fashion, i.e., simultaneously taking a small set of labeled data and a large set of unlabeled data as the training dataset to train the model. For this task, we adopt the modified Hover-Net \citep{graham2019hover}, parameterized by $\boldsymbol{\theta}$,  which is the most widely-used deep network for nuclei instance segmentation, as the base network to train our model. And we generally follow the classic Mean-Teacher network structure for semi-supervised learning.

As shown in Fig.~\ref{fig:framework}, our IRCR-Net consists of a student model $\boldsymbol{\theta}_s$ and a teacher model $\boldsymbol{\theta}_t$, both sharing an identical network structure of the modified Hover-Net \citep{graham2019hover}. For model training, each labeled image is fed into the student model to establish a supervised loss $\mathcal{L}_{sup}$. Each unlabeled image is permuted by adding noise to generate two augmented versions, a strong augmentation and a weak augmentation, which are fed into the student model and the teacher model, respectively. A consistency regularization loss $\mathcal{L}_{cons}$ is established among the activations of two augmentations. The student model $\boldsymbol{\theta}_s$ is trained via back-propagation as usual by using the total loss $\mathcal{L}_{sup} + \mathcal{L}_{cons}$, and the teacher model $\boldsymbol{\theta}_t$ is updated through an exponential moving average (EMA) mechanism \citep{tarvainen2017mean} as follows:
\begin{equation}
	\boldsymbol{\theta}_t^{(k+1)} \leftarrow \alpha \boldsymbol{\theta}_t^{(k)} + (1 - \alpha) \boldsymbol{\theta}_s^{(k+1)}, 
\end{equation}
with $\alpha = 0.95$.

The major contribution of our work is to  introduce the Instance-Aware Robust Consistency Regularization (IRCR), a novel approach to establish the consistency regularization loss, which further consists of the Matching-Driven Instance-Aware Consistency (MIAC) term and the Prior-Driven Instance-Aware Consistency (PIAC) term.

\subsection{Matching-Driven Instance-Aware Consistency}

MIAC first performs bipartite matching to align the instances predicted by the student model and the teacher model, and only those matched instances are involved in establishing the consistency loss. But one difficulty in achieving this goal is that, the outputs of Hover-Net are all continuously-valued maps which cannot distinguish individual instances. In reality, instances should be obtained by further using the watershed algorithm \citep{graham2019hover, vincent1991watersheds} over these predicted maps. To address this issue, we use the instance masks generated at the last iteration to establish the current consistency loss.  

At the $k$-th iteration, suppose the instances obtained by using the watershed algorithm from the teacher model $\mathcal{T}^{(k)} = \{\boldsymbol{T}^{(k)}_1, \boldsymbol{T}^{(k)}_2, \dots, \boldsymbol{T}^{(k)}_n\} $ and those from the student model are $\mathcal{S}^{(k)} = \{\boldsymbol{S}^{(k)}_1, \boldsymbol{S}^{(k)}_2, \dots, \boldsymbol{S}^{(k)}_m\}$. A matching between these two sets of instances can be represented by a mapping function $\sigma: \{1, 2, ..., n\} \rightarrow \{1, 2, ..., m\}$, which indicates the instance $\boldsymbol{T}^{(k)}_i \in \mathcal{T}^{(k)}$ is matched to the instance $\boldsymbol{S}^{(k)}_{\sigma(i)} \in \mathcal{S}^{(k)}$. We define the distance matrix $\boldsymbol{W} = (w_{ij})_{n \times m}$ with $w_{ij}$ being the Euclidean distance between the spatial centroid of the two instances, i.e.,
\begin{equation}
	w_{ij} = \left\|\boldsymbol{c}(\boldsymbol{T}^{(k)}_i) - \boldsymbol{c}(\boldsymbol{S}^{(k)}_j)\right\|.
\end{equation}
where $\boldsymbol{c}(\cdot)$ denotes the spatial centroid of an instance mask, computed as the mean of the pixel coordinates belonging to the instance. Given these, the task of finding an optimal matching between the two instance sets is the bipartite matching problem. And the standard Munkres algorithm \citep{kuhn1955hungarian} is utilized to obtain the mapping function $\sigma$. The MIAC loss is then defined by

\begin{equation}
	\begin{aligned}
		\label{eq:ins_loss}
		\mathcal{L}_{\text{MIAC}}^{(k+1)} = \frac{1}{N} \sum_{i=1}^{N} &\left\{ \left\| \mathbf{F}^{(k+1)}_s \odot \mathbf{S}^{(k)}_{\sigma(i)} - \mathbf{F}^{(k+1)}_t \odot \mathbf{T}^{(k)}_{i} \right\|^2\right. \\
		&+\beta \left. \left\| \mathbf{B}^{(k+1)}_s \odot \tilde{\mathbf{S}}^{(k)}_{\sigma(i)} - \mathbf{B}^{(k+1)}_t \odot \tilde{\mathbf{T}}^{(k)}_{i} \right\|^2 \right\},
	\end{aligned}
\end{equation}
where the weight $\beta$ is empirically set to $0.5$.
$\mathbf{F}$ denotes the feature maps (including both the NP and HV branches), and $\mathbf{B}$ denotes the NP-branch feature maps used for boundary emphasis. The symbol $\odot$ is the element-wise multiplication. \( \tilde{\mathbf{S}} \) and \( \tilde{\mathbf{T}} \) correspond to the extracted boundaries of matched instances \( \mathbf{S} \) and \( \mathbf{T} \) using  Sobel operator and dilation.

As highlighted by the green arrows in Fig.~\ref{fig:framework}, the instance proposals $\mathbf{S}^{(k)}$ and $\mathbf{T}^{(k)}$ are produced by a \emph{Watershed-Based Instance Segmentation} (WBIS) module from the predicted nuclear probability maps. Since WBIS contains non-differentiable operations, no gradient should flow to these masks. To address this, we adopt an iterative strategy: at iteration $k$ we (i) forward the student and teacher branches, (ii) obtain instance masks via WBIS and match them using Munkres algorithm to form matched instances, and (iii) in the next iteration $k+1$ we compute loss on the current feature maps $\big(\mathbf{F}^{(k+1)}_{s},\mathbf{B}^{(k+1)}_{s}\big)$ and
$\big(\mathbf{F}^{(k+1)}_{t},\mathbf{B}^{(k+1)}_{t}\big)$ while treating the previous instance masks $\big(\mathbf{S}^{(k)},\tilde{\mathbf{S}}^{(k)},\mathbf{T}^{(k)},\tilde{\mathbf{T}}^{(k)}\big)$ as constants.

In practice, this is equivalent to multiplying the current feature maps by the detached masks from the last stage. Unmatched instances are excluded and do not contribute to the loss, which avoids introducing noisy supervision from poor proposals.  This instance-aware consistency regularization mechanism addresses mismatches in instance quantity and position through robust instance matching, significantly reducing noise and improving the overall instance segmentation performance.

\subsection{Prior-Driven Instance-Aware Consistency}
\begin{figure}[]
	\centering
	\includegraphics[width=\linewidth]{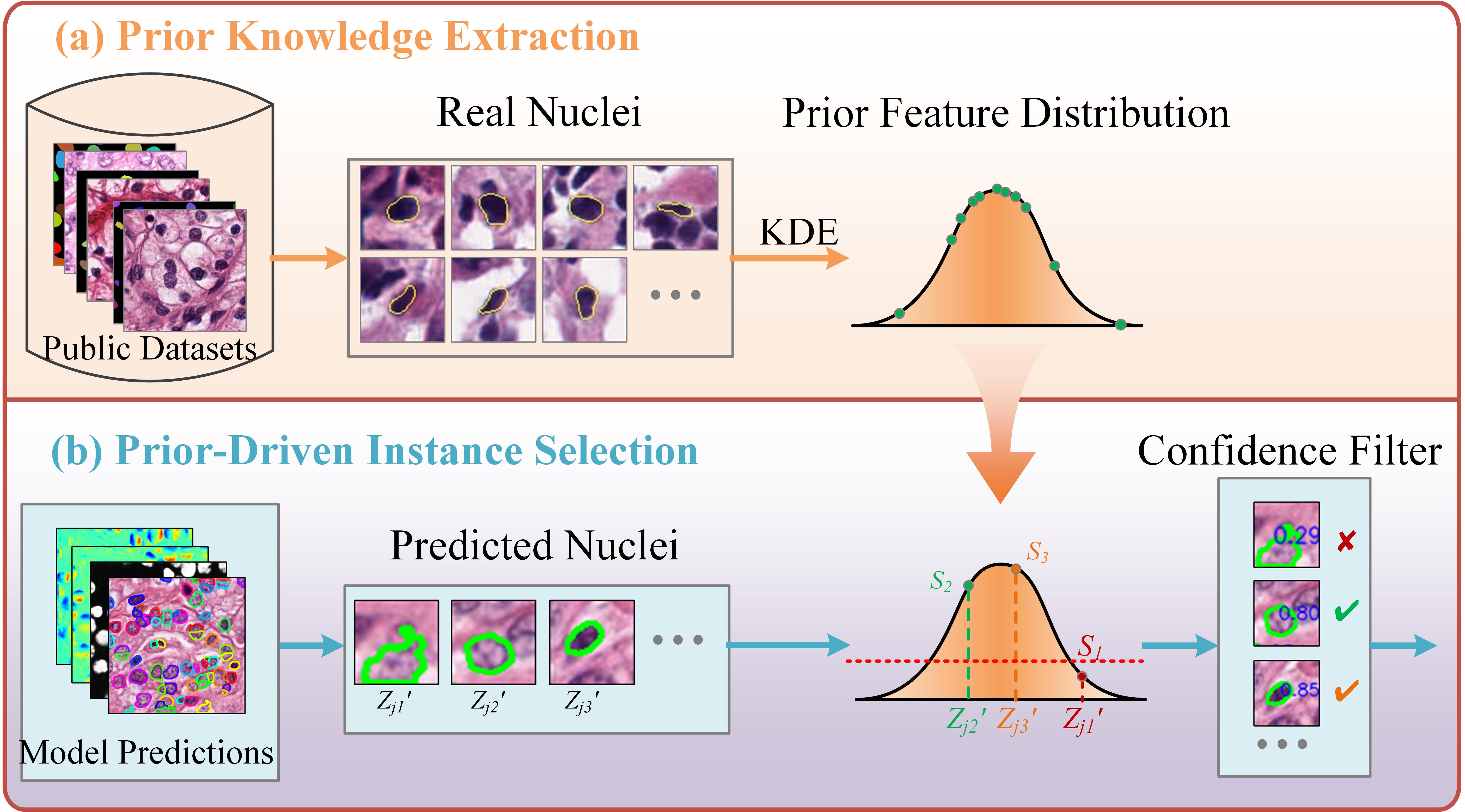} 
	\caption{Overview of the prior-driven instance selection process.}
	\label{fig:prior_instance_selection}
\end{figure}

From publicly available datasets with annotations, PIAC derives prior knowledge about nuclei  \citep{lou2024structure, sharma2015multi}, which is represented in terms of the statistical distributions of certain handcrafted features, as detailed in Table \ref{tab:nucleus_features} and Fig~\ref{fig:prior_instance_selection}. We extract features from publicly available datasets $\mathcal{D}_{ext}$ (e.g., MoNuSAC \citep{verma2021monusac2020}, CoNSeP \citep{graham2019hover}, PanNuke \citep{gamper2019pannuke}, and Lizard \citep{graham2021lizard}), ensuring no overlap with the target dataset $\mathcal{D}$, i.e., $\mathcal{D}_{ext} \cap \mathcal{D} = \phi$. Since the forms of distributions are unknown, we utilize the non-parametric kernel density estimation (KDE) approach to calculate the distributions.

Formally, over the feature channel $x_1$, suppose the features extracted from the training samples are denoted by $\mathcal{X}_1 = \{x_1^{(n)}\}_{n=1}^N$. The prior distribution is defined by
\begin{equation}
	\label{eq:pdf}
	p(x_1) = \frac{1}{\sqrt{2\pi} N h} \sum_{n=1}^{N} \exp\left[ -\frac{\left(x_1 - x_1^{(n)} \right)^2}{2h^2}\right],
\end{equation}
where $h$ is the bandwidth that controls the smoothness of the density estimate. The distributions over the other feature channels $p(x_i)$ ($i = 2, 3, 4, \dots$) can be calculated in the same way. Then, at the inference stage, given a sample represented by $\boldsymbol{z} = (z_1, z_2, ..., z_K)^{\mathsf T}$, its likelihood of being a real nucleus is evaluated by
\begin{equation}
	p(\boldsymbol{z}) = \frac{1}{K} \sum_{k=1}^K p(z_k).
\end{equation}
Given these, at the $k$-th iteration, every instance $\boldsymbol{T}^{(k)}_j$ can be assigned with a score $p(\boldsymbol{z}_j)$, based on which a mask can be defined by
\begin{equation}
	\small
	\mathbf{U}^{(k)}(x, y) =
	\begin{cases}
		0, & \text{if } (x, y) \in \boldsymbol{T}^{(k)}_j, p(\boldsymbol{z}_j) < \tau \\ 
		w, & \text{otherwise}
	\end{cases} ,
\end{equation}
where \( \tau \) represents the distance threshold, which is empirically set to 0.35. \( w \) is a weighting factor that amplifies the contribution of reliable pseudo-labels (\( w \) = 2), ensuring that high-quality predictions exert a stronger influence during training.

The PIAC loss is then defined by 

\begin{equation}
	\begin{aligned}
		\label{eq:PIAC_loss}
		\mathcal{L}_{\text{PIAC}}^{(k+1)} = \frac{1}{N} \sum_{i=1}^{N}  \left\| \left( \mathbf{F}^{(k+1)}_s - \mathbf{F}^{(k+1)}_t \right) \odot \mathbf{U}^{(k)}_{i} \right\|^2.
	\end{aligned}
\end{equation}

\subsection{Total Loss Function}

For labeled pathological image data \((X_l, Y_l)\), the student model \(\theta_s\) generates two branch predictions \(\hat{\mathbf{N}}\) and \(\hat{\mathbf{HV}}\) from the input image \(X_l\). For the NP branch, this work employs the Dice loss \(\mathcal{L}_{Dice}\) and cross-entropy (CE) loss \(\mathcal{L}_{CE}\) \citep{graham2019hover}, as follows:
\begin{equation}
	\vspace{-5pt}
	\small
	\mathcal{L}_{Dice} = 1 - \frac{2 \sum_{i=1}^N \hat{y}_i \cdot y_i + \epsilon}{\sum_{i=1}^N \hat{y}_i + \sum_{i=1}^N y_i + \epsilon},
\end{equation}
\begin{equation}
	\small
	\mathcal{L}_{CE} = -\frac{1}{N} \sum_{i=1}^N \sum_{c=1}^C y_{i,c} \log \hat{y}_{i,c},
	\vspace{-5pt}
\end{equation}
where \(\epsilon = e^{-3}\) is a smoothing constant to avoid division by zero, \(C\) denotes the number of classes (\(C\) = 2). For the HV branch, this work employs mean squared error (MSE) loss \(\mathcal{L}_{MSE}\) and mean squared gradient error (MSGE) loss \(\mathcal{L}_{MSGE}\) \citep{graham2019hover} to constrain the predicted distance maps, enhancing the recognition of overlapping and blurred boundaries, as follows:
\begin{equation}
	\mathcal{L}_{MSE}=\frac{1}{n} \sum_{i=1}^n\left(y_i-\hat{y}_i\right)^2,
\end{equation}
\begin{equation}
	\small
	\mathcal{L}_{MSGE} = \frac{1}{m} \sum_{i \in M} \left[ (\nabla_x \hat{y}_i - \nabla_x y_i)^2 + (\nabla_y \hat{y}_i - \nabla_y y_i)^2 \right],
\end{equation}
where  \(n\) is the total number of pixels in the feature map, \(\nabla_x\) and \(\nabla_y\) are the horizontal and vertical gradients. \(M\) denotes the set of all pixels belonging to nuclear regions, and \(m\) is the total number of such nuclear pixels within the image. Finally, the supervised loss \(\mathcal{L}_{sup}\) is formulated as:
\begin{equation}
	\small
	\begin{aligned}
		\mathcal{L}_{sup} &= 
		\underbrace{ \mathcal{L}_{Dice}\left(\hat{\mathbf{N}}, \mathbf{N}_{GT}\right) 
			+  \mathcal{L}_{CE}\left(\hat{\mathbf{N}}, \mathbf{N}_{GT}\right)}_{\text{the NP branch term}} \\
		& \quad + 
		\underbrace{ \mathcal{L}_{MSE}\left(\hat{\mathbf{HV}}, \mathbf{HV}_{GT}\right)
			+  \mathcal{L}_{MSGE}\left(\hat{\mathbf{HV}}, \mathbf{HV}_{GT}\right)}_{\text{the HV branch term}},
	\end{aligned}
\end{equation}
where \( \hat{\mathbf{N}} \) and \( \hat{\mathbf{HV}} \) denote the predicted feature maps of the NP branch and HV branch, respectively, while \( \mathbf{N}_{GT} \) and \( \mathbf{HV}_{GT} \) represent the corresponding ground truth. The NP branch term ensures accurate nuclear foreground-background segmentation, while the HV branch term enhances the separation of touching and overlapping nuclei by constraining the predicted horizontal and vertical distance maps.

\begin{table}[]
	\centering
	\small
	\caption{Descriptions and formulas for morphological features of nuclei.}
	\begin{tabular}{@{}>{\centering\arraybackslash}m{1.0cm}
			>{\centering\arraybackslash}m{4.2cm}
			>{\centering\arraybackslash}m{3.0cm}@{}}
		
		\toprule
		Feature       & Description                     & Formula \\ 
		\midrule
		Area                  & Nucleus area                               & \(z_1^{\prime}=S_{\text {area }}\) \\ 
		Solidity              & Ratio of area to convex hull               & \(z_2^{\prime}=S_{\text{area}} / S_{\text{hull}}\) \\ 
		Circularity           & Shape circularity                          & \(z_3^{\prime}=4\pi S_{\text{area}} / \text{perimeter}^2\) \\ 
		Intensity             & Average intensity in $H$ channel              & \(z_4^{\prime}=\frac{1}{N} \sum_{i=1}^N H_i\) \\ 
		Extent                & Ratio of area to bounding rectangle        & \(z_5^{\prime}=S_{\text{area}} / S_{\text{BR}}\) \\ 
		\bottomrule
	\end{tabular}
	\label{tab:nucleus_features}
\end{table}

For unlabeled data \(X_u\), the pseudo-labels generated by the teacher model are filtered and weighted using the prior-driven quality assessment and instance-aware consistency regularization. The consistency loss \(\mathcal{L}_{cons}\) for unlabeled data is expressed as:
\begin{equation}
	\vspace{-2pt}
	\small
	\mathcal{L}_{cons}=\gamma_1 \mathcal{L}_{PIAC}+\gamma_2 \mathcal{L}_{MIAC},
	\vspace{-2pt}
\end{equation}
where the balancing weights \(\gamma_1\) and \(\gamma_2\) are set to 0.1 and 100 based on experimental observations.

The total loss \(\mathcal{L}\) is the sum of the supervised loss \(\mathcal{L}_{sup}\) and the consistency loss \(\mathcal{L}_{cons}\), defined as:
\begin{equation}
	\vspace{-2pt}
	\small
	\mathcal{L} = \mathcal{L}_{sup} + \mathcal{L}_{cons}.
	\vspace{-2pt}
\end{equation}
This comprehensive loss function, incorporating both supervised and unsupervised constraints, effectively addresses the scarcity of labeled data, significantly improves pseudo-label reliability, and ensures instance-level consistency in semi-supervised nuclei instance segmentation (details in Table \ref{table:loss_ablation}).

\begin{table*}[!t]
	\centering
	\setlength{\tabcolsep}{4pt}
	\caption{Comparison of segmentation performance between semi-supervised and supervised methods on the MoNuSeg, MoNuSAC, PanNuke, and CoNSeP datasets under varying labeled training data ratios.}
	\label{table:comparison_SOTA}
	\renewcommand{\arraystretch}{1.1} 
	\small
	\begin{tabular}{cccccccccccccc}
		\toprule
		\multirow{2}{*}{Label} & \multirow{2}{*}{Method} & \multicolumn{3}{c}{MoNuSeg}                      & \multicolumn{3}{c}{MoNuSAC}                      & \multicolumn{3}{c}{PanNuke}                      & \multicolumn{3}{c}{ConSep}                       \\ \cline{3-14} 
		&                         & AJI            & Dice           & $\mathrm{F1}_{obj}$           & AJI            & Dice           & $\mathrm{F1}_{obj}$           & AJI            & Dice           & $\mathrm{F1}_{obj}$           & AJI            & Dice           & $\mathrm{F1}_{obj}$           \\ \hline
		\multirow{6}{*}{1/32}  & Hover-Net\citep{graham2019hover}               & 0.478          & 0.691          & 0.732          & 0.471          & 0.664          & 0.721          & 0.542          & 0.745          & 0.731          & 0.353          & 0.690          & 0.598          \\
		& ST\citep{zhu2021improving}                     & 0.442          & 0.653          & 0.694          & 0.451          & 0.625          & 0.711          & 0.475          & 0.670          & 0.678          & 0.341          & 0.629          & 0.553          \\
		& MT\citep{tarvainen2017mean}                      & 0.573          & 0.767          & 0.804          & 0.510          & 0.687          & 0.754          & 0.574          & 0.756          & 0.763          & 0.446          & 0.747          & 0.663          \\
		& CDCL\citep{wu2022cross}                    & -              & 0.732          & -              & -              & 0.660          & -              & -              & 0.677          & -              & -              & 0.694          & -              \\
		& PG-FANet\citep{jin2024inter}                & 0.460          & 0.669          & 0.779          & 0.370          & 0.616          & 0.712          & 0.381          & 0.657          & 0.638          & 0.218          & 0.641          & 0.568          \\
		& IRCR-Net(Ours)             & \textbf{0.591} & \textbf{0.780} & \textbf{0.805} & \textbf{0.537} & \textbf{0.719} & \textbf{0.766} & \textbf{0.576} & \textbf{0.766} & 0.762          & \textbf{0.450} & \textbf{0.748} & \textbf{0.666} \\ \hline
		\multirow{6}{*}{1/16}  & Hover-Net\citep{graham2019hover}                & 0.586          & 0.755          & 0.789          & 0.515          & 0.691          & 0.748          & 0.590          & 0.771          & 0.763          & 0.428          & 0.754          & 0.650          \\
		& ST\citep{zhu2021improving}                      & 0.470          & 0.697          & 0.722          & 0.488          & 0.655          & 0.738          & 0.544          & 0.738          & 0.732          & 0.419          & 0.745          & 0.610          \\
		& MT\citep{tarvainen2017mean}                      & 0.613          & 0.793          & 0.825          & 0.539          & 0.707          & 0.775          & 0.603          & 0.774          & 0.786          & 0.483          & 0.773          & 0.697          \\
		& CDCL\citep{wu2022cross}                    & -              & 0.758          & -              & -              & 0.701          & -              & -              & 0.721          & -              & -              & 0.713          & -              \\
		& PG-FANet\citep{jin2024inter}                & 0.489          & 0.786          & 0.745          & 0.385          & 0.689          & 0.718          & 0.460          & 0.707          & 0.667          & 0.266          & 0.709          & 0.580          \\
		& IRCR-Net(Ours)             & \textbf{0.614} & 0.791          & 0.821          & \textbf{0.562} & \textbf{0.731} & \textbf{0.784} & \textbf{0.609} & \textbf{0.782} & \textbf{0.789} & \textbf{0.490} & \textbf{0.777} & \textbf{0.697} \\ \hline
		\multirow{6}{*}{1/8}   & Hover-Net\citep{graham2019hover}                & 0.544          & 0.734          & 0.786          & 0.537          & 0.706          & 0.767          & 0.616          & 0.786          & 0.780          & 0.488          & 0.785          & 0.699          \\
		& ST\citep{zhu2021improving}                      & 0.443          & 0.653          & 0.673          & 0.511          & 0.677          & 0.755          & 0.564          & 0.747          & 0.734          & 0.427          & 0.746          & 0.601          \\
		& MT\citep{tarvainen2017mean}                      & 0.625          & 0.797          & 0.835          & 0.546          & 0.710          & 0.782          & 0.624          & 0.787          & 0.796          & 0.516          & 0.793          & 0.720          \\
		& CDCL\citep{wu2022cross}                    & -              & 0.774          & -              & -              & 0.729          & -              & -              & 0.753          & -              & -              & 0.744          & -              \\
		& PG-FANet\citep{jin2024inter}                & 0.501          & 0.787          & 0.815          & 0.386          & 0.702          & 0.719          & 0.402          & 0.729          & 0.693          & 0.261          & 0.730          & 0.583          \\
		& IRCR-Net(Ours)             & \textbf{0.633} & \textbf{0.801} & \textbf{0.836} & \textbf{0.570} & \textbf{0.735} & \textbf{0.793} & \textbf{0.631} & \textbf{0.795} & \textbf{0.804} & \textbf{0.521} & \textbf{0.793} & \textbf{0.725} \\ \hline
		\multirow{6}{*}{1/4}   & Hover-Net\citep{graham2019hover}                & 0.601          & 0.775          & 0.815          & 0.565          & 0.728          & 0.793          & 0.643          & 0.805          & 0.802          & 0.516          & 0.793          & 0.718          \\
		& ST\citep{zhu2021improving}                      & 0.557          & 0.774          & 0.754          & 0.553          & 0.719          & 0.781          & 0.596          & 0.781          & 0.754          & 0.458          & 0.745          & 0.645          \\
		& MT\citep{tarvainen2017mean}                      & 0.627          & 0.795          & 0.836          & 0.554          & 0.718          & 0.783          & 0.645          & 0.802          & 0.812          & 0.538          & 0.805          & 0.738          \\
		& CDCL\citep{wu2022cross}                    & -              & 0.789          & -              & -              & 0.750          & -              & -              & 0.775          & -              & -              & 0.770          & -              \\
		& PG-FANet\citep{jin2024inter}                & 0.526          & 0.797          & 0.827          & 0.433          & 0.718          & 0.759          & 0.413          & 0.732          & 0.701          & 0.246          & 0.725          & 0.560          \\
		& IRCR-Net(Ours)             & \textbf{0.641} & \textbf{0.805} & \textbf{0.838} & \textbf{0.574} & 0.737          & \textbf{0.793} & \textbf{0.650} & \textbf{0.807} & \textbf{0.816} & \textbf{0.541} & 0.803          & \textbf{0.738} \\ \hline
		100\%                  & FullSup\citep{graham2019hover}                 & 0.617          & 0.780          & 0.831          & 0.565          & 0.728          & 0.793          & 0.664          & 0.817          & 0.820          & 0.543          & 0.804          & 0.742          \\ 
\bottomrule
	\end{tabular}
\end{table*}

\section{Experiments}

\subsection{Dataset and Evaluation Metrics}
\subsubsection{Datasets}
We evaluated the proposed method on four public nuclei instance segmentation datasets: MoNuSeg \citep{kumar2019multi}, MoNuSAC \citep{verma2021monusac2020}, PanNuke \citep{gamper2019pannuke}, and CoNSeP \citep{graham2019hover}.

\paragraph{MoNuSeg} The MoNuSeg dataset consists of $1000 \times 1000$ pixel patches extracted from whole-slide images (WSIs) of seven organs (e.g., breast, liver, kidney) in The Cancer Genome Atlas program (TCGA), scanned at $40\times$ magnification. It includes 21,623 manually annotated nuclei across 30 training images and 14 testing images. In our experiments, we randomly sampled 90\% of the training set for model training, reserving the remaining 10\% for validation.

\paragraph{MoNuSAC} The MoNuSAC dataset, introduced in the ISBI 2020 challenge, consists of variable-sized images sourced from the TCGA database, scanned at $40\times$ magnification. It comprises 209 training images and 101 testing images. We randomly selected 90\% of the training images for model training, while reserving the remaining 10\% for validation.

\paragraph{PanNuke} The PanNuke dataset consists of $256 \times 256$ patches from 19 organs, with 216.4K nuclei instances semi-automatically annotated. Following the official split \citep{gamper2019pannuke}, we use Fold 1 (2656 images) as the training set, Fold 2 (2523 images) as the validation set, and Fold 3 (2722 images) as the test set.

\paragraph{CoNSeP} The CoNSeP dataset consists of 41 H\&E-stained $1000 \times 1000$ patches from colorectal adenocarcinomas, extracted from 16 WSIs scanned at $40\times$ magnification. The dataset includes 27 images for training and 14 for testing, covering diverse pathological contexts. We split the training set by randomly selecting 90\% for training and the remaining 10\% for validation.

We cropped all images into non-overlapping $256 \times 256$ patches during preprocessing. Experiments were conducted under four labeled data ratios: $1/32$, $1/16$, $1/8$, and $1/4$ to evaluate the semi-supervised framework comprehensively. In each setting, a corresponding proportion of labeled data was randomly sampled for training, while the rest was treated as unlabeled. Each experiment was repeated with different random seeds three times, and the average performance was reported as the final result.

\subsubsection{Evaluation Metrics}
To evaluate the segmentation accuracy of nuclei instances, this study adopts widely used metrics for nuclei instance segmentation, including the Aggregated Jaccard Index (AJI) \citep{kumar2017dataset}, the Dice coefficient (Dice) \citep{taha2015metrics}, and the object-level F1-score ($\mathrm{F1} _{obj}$) \citep{zhang2024venet}.

\paragraph{Aggregated Jaccard Index (AJI)} 
The AJI metric considers both pixel-level segmentation accuracy and instance-level matching, making it more suitable for evaluating instance segmentation tasks than pure pixel-based Intersection over Union (IoU). The formula is defined as:

\begin{equation}
	\mathrm{AJI} = \frac{\sum_{i=1}^{N} G_i \cap S_j}{\sum_{i=1}^{N} G_i \cup S_j + \sum_{S_k \in U} S_k},
\end{equation}
where \( G_i \) represents the connected region of the \( i \)-th nucleus in the ground truth, \( S_j \) is the predicted instance mask that has the maximum IoU with \( G_i \), and \( S_k \) denotes the predicted masks that are unmatched with any \( G_i \).

\paragraph{Dice Coefficient} 
Dice is a pixel-based evaluation metric widely applied in segmentation and classification tasks. Its formula is as follows:

\begin{equation}
	\mathrm{Dice} = \frac{2 |G \cap S|}{|G| + |S|},
\end{equation}
where \( G \) is the set of pixels in the ground truth, and \( S \) is the set of pixels in the predicted segmentation mask.

\paragraph{${F1} _{obj}$} 
The object-level F1-score ($\mathrm{F1} _{obj}$) is a general metric for evaluating instance detection performance, determined by true positives (TP), false positives (FP), and false negatives (FN). Its formula is expressed as:
\begin{equation}
	\mathrm{F1} _{obj} = \frac{2 \mathrm{TP}}{2 \mathrm{TP} + \mathrm{FP} + \mathrm{FN}}.
\end{equation}

\subsection{Implementation Details}
We conducted the experiments on a server equipped with four NVIDIA RTX 3090 GPUs. The proposed method was implemented in PyTorch, utilizing the Adam optimizer with a batch size of 4 and an initial learning rate of $1 \times 10^{-4}$. For the MoNuSeg and CoNSeP datasets, the model was trained for 150 epochs, with the learning rate reduced to 10\% of its initial value after 100 epochs. For the MoNuSAC and PanNuke datasets, training was conducted for 50 epochs, with the learning rate reduced by a factor of 0.1 every 25 epochs. 

Data augmentation included random flips, scaling, rotation, brightness, contrast, saturation adjustments, Gaussian noise addition, and center cropping. All images were cropped to $256 \times 256$ to maintain uniform input dimensions. Models were initialized with ImageNet pre-trained weights. The final evaluation model was obtained from the student network at the last training epoch.

\subsection{Comparisons with Other Methods}
\begin{figure}
	\centering
	\setlength{\belowcaptionskip}{-0.2cm}
	\includegraphics[scale = 0.75]{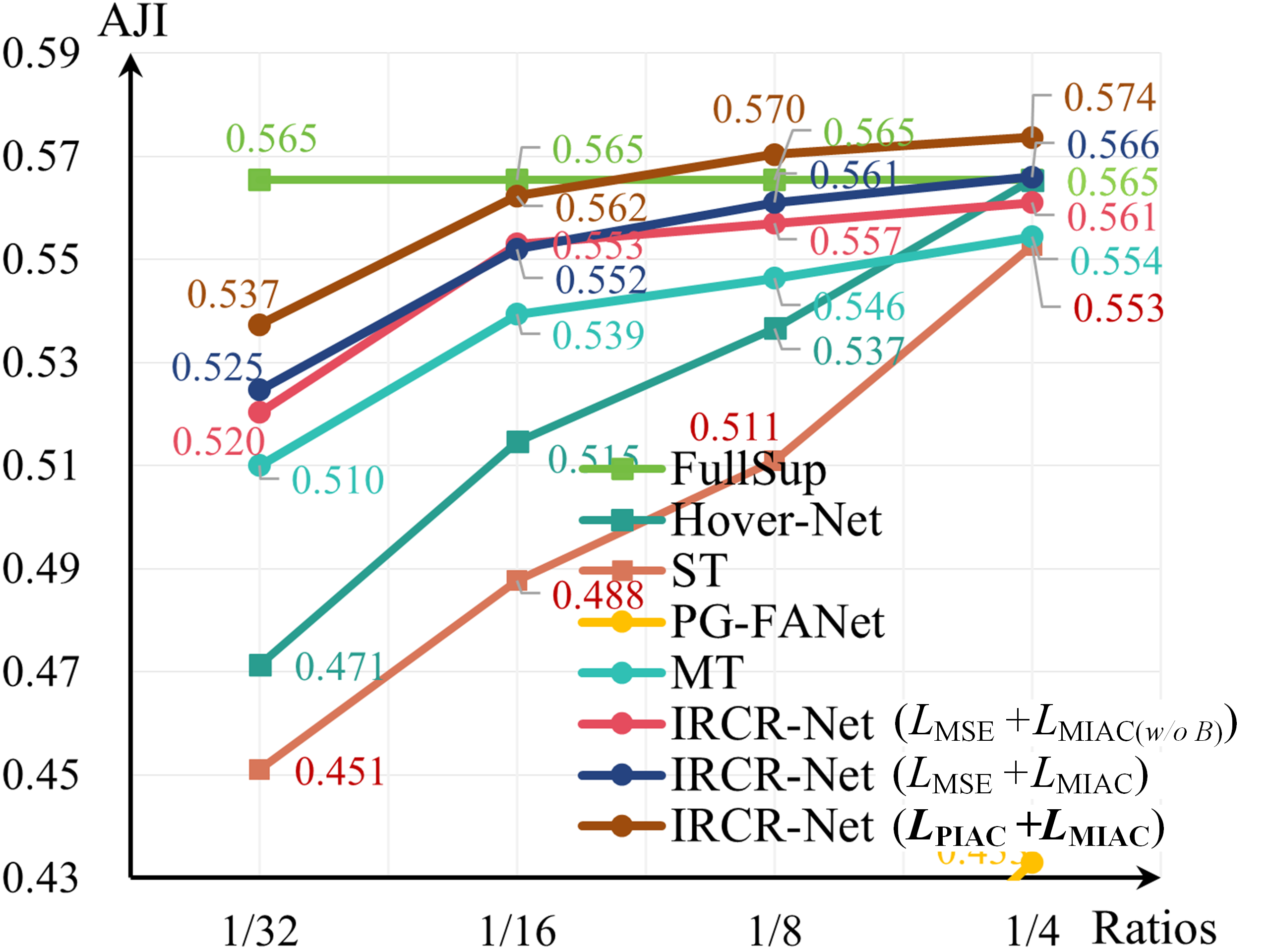}
	\caption{Comparison with other methods on MoNuSAC datasets under varying labeled training data ratios.}
	\label{fig:viz_result}
\end{figure}

To evaluate the effectiveness of the proposed IRCR-Net, we compared it with several other methods: Hover-Net \citep{graham2019hover}, Self-Training (ST) \citep{zhu2021improving}, Mean-Teachers (MT) \citep{tarvainen2017mean}, CDCL \citep{wu2022cross}, PG-FANet \citep{jin2024inter} and fully supervised segmentation (FulSup) methods based on Hover-Net. The results on MoNuSeg, MoNuSAC, PanNuke, and CoNSeP datasets under varying labeled data ratios (1/32, 1/16, 1/8, 1/4) are shown in Table \ref{table:comparison_SOTA} and Fig. \ref{fig:viz_result}. Our method outperforms competing methods, particularly under highly limited labeled data conditions (e.g., 1/32).

\begin{figure*}
	\centering
	\setlength{\belowcaptionskip}{-0.2cm}
	\includegraphics[scale = 0.92]{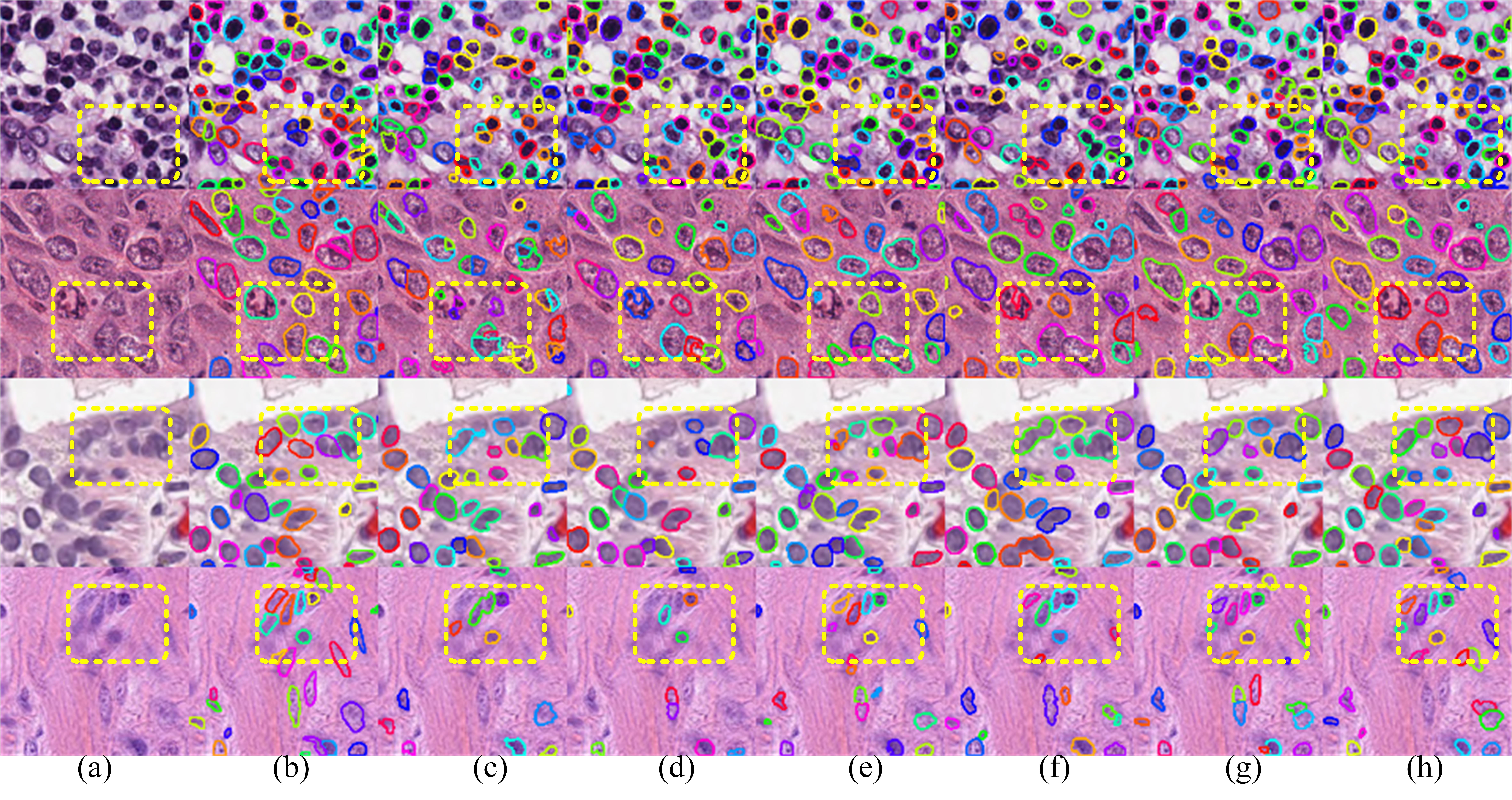}
	
	\caption{Visual comparison of nuclei segmentation results from different methods: (a) Input image, (b) Ground truth, (c) Hover-Net, (d) ST, (e) MT, (f) PG-FANet, (g) Ours, and (h) FullSup.}
	\label{fig:viz_inst_images}
\end{figure*}

Hover-Net, a classical supervised approach for nuclei instance segmentation, suffers from performance degradation due to limited labeled data. Under the 1/32 and 1/16 labeled data configurations, its segmentation accuracy drops significantly compared to the fully supervised method (FullSup). For example, on the CoNSeP dataset, the AJI metric declines from 0.543 (FullSup) to 0.353 (1/32). ST (Self-Training) method first trains the model using the ground truth labels and progressively replaces the original labels with their own pseudo-labels during training. However, this unconstrained pseudo-labeling approach proves detrimental. This decline is attributed to the lack of an effective pseudo-label quality control mechanism, which introduces erroneous pseudo-labels into the training process. These low-quality pseudo-labels reinforce biases in self-training iterations, ultimately degrading model performance. MT method leverages consistency constraints to improve performance over Hover-Net and ST. For instance, under 1/32 labeled data on MoNuSAC, MT achieves an AJI of 0.510, outperforming Hover-Net (0.471) but still falling short of our method (0.537). CDCL is a SOTA method focusing on overall pixel-wise nuclei semantic segmentation rather than instance-level metrics. Therefore, AJI and $\mathrm{F1} _{obj}$ metrics are not reported for CDCL. Its Dice scores are still lower than our method, and the lack of instance-level segmentation capabilities limits its application for tasks requiring precise instance delineation. PG-FANet achieves relatively high Dice scores but exhibits lower AJI values. While it effectively segments nuclear clusters, it struggles to separate individual touching nuclei accurately (as shown in Fig. \ref{fig:viz_inst_images}(f)), limiting its instance segmentation performance. 

The proposed IRCR-Net not only demonstrates superior performance under low annotation settings but also has the potential to surpass fully supervised models (FullSup) when more labeled data is available. For example, under the 1/4 labeled data configuration, IRCR-Net achieves higher AJI, Dice and $\mathrm{F1}_{obj}$ scores than the fully supervised approach on both the MoNuSeg and MoNuSAC datasets. It indicates that IRCR-Net effectively utilizes unlabeled data, mitigating performance degradation caused by limited labeled samples.

In summary, our IRCR-Net achieves superior performance across various datasets and labeling configurations. It demonstrates strong robustness in low-label scenarios, underscoring its potential for semi-supervised nuclei instance segmentation.

\begin{table}[t]
	\centering
		\small  
	\caption{Ablation study of loss function components on the MoNuSAC dataset with 1/32 labeled training data configuration. \textit{w/o B} means without instance boundary enhanced consistency calculation.}
	\label{table:loss_ablation}
	\setlength{\tabcolsep}{2.5pt}  
	\renewcommand{\arraystretch}{1.1}  
	
	\begin{tabular}{ccccccccc}
		\toprule
		Method   & $\mathcal{L}_{\text{sup}}$ & $\mathcal{L}_{\text{MSE}}$ & $\mathcal{L}_{\text{PIAC}}$ & $\mathcal{L}_{\text{MIAC(\textit{w/o B})}}$ & $\mathcal{L}_{\text{MIAC}}$ & AJI & Dice & $\mathrm{F1}_{\textit{obj}}$ \\
		\midrule
		\textit{SupOnly}   & \checkmark &       &        &        &        & 0.471 & 0.664 & 0.721 \\
		\textit{Scheme.1}  & \checkmark & \checkmark &        &        &        & 0.510 & 0.687 & 0.754 \\
		\textit{Scheme.2}  & \checkmark & \checkmark &        & \checkmark       &  & 0.520 & 0.697 & 0.754 \\
		\textit{Scheme.3}  & \checkmark & \checkmark &        &  & \checkmark       & 0.525 & 0.701 & 0.765 \\
		\textit{Ours}      & \checkmark &          & \checkmark &  & \checkmark & \textbf{0.537} & \textbf{0.719} & \textbf{0.766} \\
		\bottomrule
	\end{tabular}
\end{table}

\begin{table}[t]
	\centering
	\small
	\caption{Effectiveness of instance matching on segmentation performance on the MoNuSAC Datasets with 1/32 labeled training data configuration.}
	\label{table:instance_pairing}
	\renewcommand{\arraystretch}{1.1} 
	\begin{tabular}{ccccc}
		\toprule
		Method                                 & \textit{r} & AJI            & Dice           & $\mathrm{F1}_{obj}$             \\ \hline
		\textit{w/o Matching}                  & -          & 0.510          & 0.687          & 0.754          \\ \hline
		\multirow{4}{*}{\textit{w   Matching}} & 0.1        & 0.516          & 0.694          & 0.757          \\
		& 1.0        & 0.524          & 0.700          & 0.764          \\
		& 1.5        & \textbf{0.525} & \textbf{0.701} & \textbf{0.765} \\
		& 2.0        & 0.524          & 0.700          & 0.764          \\
		& 3.0        & 0.522          & 0.696          & 0.763          \\ \bottomrule
	\end{tabular}
\end{table}

\subsection{Ablation Experiments and Analysis}

\subsubsection{Effectiveness of Loss Functions}
We conducted ablation studies on loss functions to evaluate the contributions of each component, as shown in Table \ref{table:loss_ablation} and Fig. \ref{fig:viz_result}. Under a labeled training data ratio of 1/32, applying only the supervised loss (\textit{SupOnly}) resulted in a low AJI of 0.471, highlighting the performance limitations caused by insufficient labeled data. Introducing the MSE-based consistency loss (\textit{Scheme. 1}) improved the performance, but the limited effectiveness of global consistency constraints in complex pathological image segmentation still needs to be improved. Incorporating the $\mathcal{L}_{\text{MIAC(\textit{w/o B})}}$ without instance boundary enhanced consistency calculation (\textit{Scheme. 2}), which strengthens instance-level feature learning and improves boundary delineation, improved AJI to 0.520, highlighting the importance of instance-specific mechanisms. The inclusion of the boundary-enhanced consistency loss $\mathcal{L}_{\text{MIAC}}$ (\textit{Scheme. 3}) further increased AJI, Dice and $\mathrm{F1}_{obj}$, demonstrating the benefits of fine-grained boundary attention for resolving overlapping nuclei.

The complete model (\textit{Ours}) refined $\mathcal{L}_{MSE}$ into prior knowledge-guided consistency loss $\mathcal{L}_{PIAC}$, combined with $\mathcal{L}_{MIAC}$ (matching-driven instance-aware consistency with instance boundary enhanced), achieving the highest AJI of 0.537, Dice of 0.719, and $\mathrm{F1}_{obj}$ of 0.766. These results validate the effectiveness of the proposed consistency loss components, which collectively improve nuclei instance segmentation performance.

\subsubsection{Effectiveness of Instance Matching}
We conducted ablation studies to assess the impact of the instance matching mechanism on nuclei segmentation performance. As shown in Table \ref{table:instance_pairing}, incorporating instance matching (\textit{w/ Matching}) significantly improves segmentation accuracy compared to models without it (\textit{w/o Matching}). The instance matching mechanism establishes one-to-one correspondences between teacher and student model predictions, ensuring that consistency loss is applied only to well-aligned instances, thereby mitigating errors caused by mismatched predictions. A small distance threshold \(r\) (e.g., \(r=0.1\)) limits the number of valid matches, weakening the effectiveness of consistency constraints. Conversely, an excessively large \(r\) may reduce segmentation accuracy. However, it still outperforms the model without instance matching. The optimal threshold is found at \(r=1.5\), highlighting the importance of appropriately balancing match precision and instance coverage to maximize performance in semi-supervised nuclei segmentation.

\begin{figure*}
	\centering
	\setlength{\belowcaptionskip}{-0.2cm}
	\includegraphics[scale = 0.9]{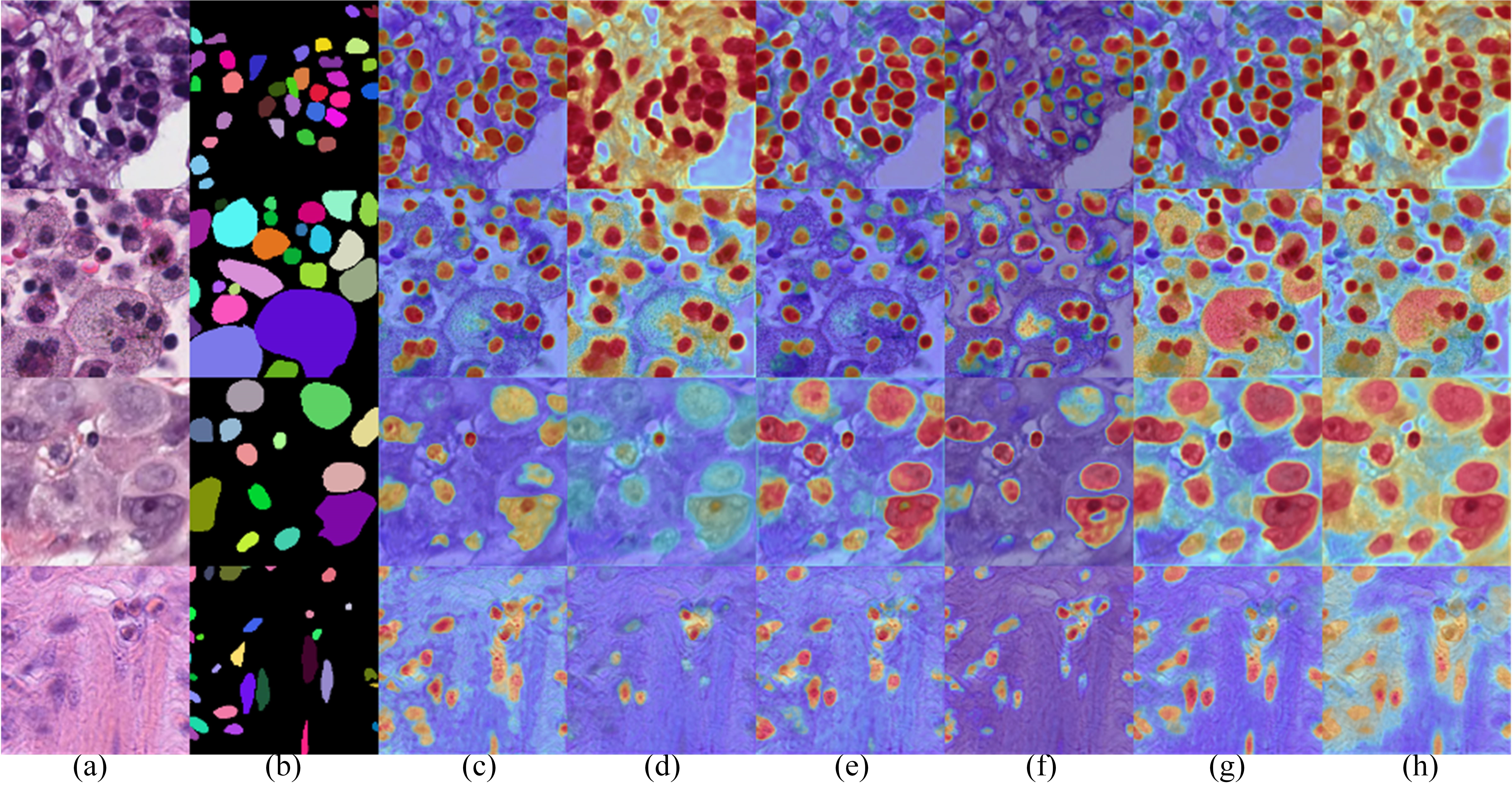}
	\caption{Visual comparison of feature maps extracted from different methods: (a) Input image, (b) Ground truth, (c) Hover-Net, (d) ST, (e) MT, (f) PG-FANet, (g) Ours, and (h) FullSup.}
	\label{fig:viz_att_images}
\end{figure*}

\subsection{Visualization Results}
We visualized different methods' segmentation results and feature attention maps, as shown in Fig. \ref{fig:viz_inst_images} and Fig. \ref{fig:viz_att_images}. Our proposed method achieves superior performance in accurately segmenting overlapping nuclei and delineating intricate boundaries, nearing the results of fully supervised models. The feature attention maps provide deeper interpretability into the observed improvements. For example, the Hover-Net and PG-FANet methods display scattered and less focused attention distributions, highlighting their limited ability to extract salient features from target regions. In contrast, our method effectively concentrates attention on nuclei regions, emphasizing key structures and boundaries. 

\begin{figure}
	\centering
	\setlength{\belowcaptionskip}{-0.2cm}
	\includegraphics[scale = 0.35]{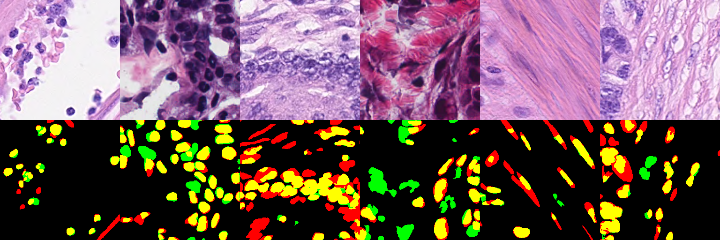}
	\caption{Failure cases for our method. The first row shows the original input images, and the second shows failure cases. The green pixels represent predictions, the red pixels indicate ground truth and the yellow pixels denote their overlapping regions.}
	\label{fig:viz_faliure_cases}
\end{figure}

The failure cases illustrated in Fig. \ref{fig:viz_faliure_cases} highlight limitations in handling highly overlapping nuclei, indistinct boundaries, and regions with significant morphological variability. While our method demonstrates strong robustness in four public datasets, these cases point to areas for refinement. Future efforts will focus on refining the model's ability to handle extreme complexities, such as ambiguous boundaries and densely packed nuclei, to enhance its applicability to real-world pathological scenarios.

\section{Conclusion}
The proposed IRCR-Net effectively integrates instance-level prior knowledge and instance-aware consistency loss designs to filter low-quality predictions, significantly reducing the impact of label noise on model training. This approach achieves remarkable performance improvements in semi-supervised nuclei instance segmentation under limited annotated data conditions, reaching or even surpassing the performance of fully supervised methods. These advancements lay a solid foundation for subsequent applications in pathological image analysis and tumor microenvironment research. However, our method still exhibits limitations in highly challenging scenarios, such as densely overlapping or blurred-boundary nuclei. Future work will focus on incorporating more advanced multi-task and multi-modal learning strategies to enhance the framework's generalizability and adaptability for other instance segmentation tasks.

\section*{Acknowledgments}
This work was funded by Guangdong Basic and Applied Basic Research Foundation under Grant 2024A1515010258 and Grant 2025A1515010250.

\bibliographystyle{model2-names.bst}\biboptions{authoryear}
\bibliography{references}

\begin{thebibliography}{50}
\expandafter\ifx\csname natexlab\endcsname\relax\def\natexlab#1{#1}\fi
\providecommand{\url}[1]{\texttt{#1}}
\providecommand{\href}[2]{#2}
\providecommand{\path}[1]{#1}
\providecommand{\DOIprefix}{doi:}
\providecommand{\ArXivprefix}{arXiv:}
\providecommand{\URLprefix}{URL: }
\providecommand{\Pubmedprefix}{pmid:}
\providecommand{\doi}[1]{\href{http://dx.doi.org/#1}{\path{#1}}}
\providecommand{\Pubmed}[1]{\href{pmid:#1}{\path{#1}}}
\providecommand{\bibinfo}[2]{#2}
\ifx\xfnm\relax \def\xfnm[#1]{\unskip,\space#1}\fi
\bibitem[{Bilodeau et~al.(2022)Bilodeau, Delmas, Parent, De~Koninck, Durand and
  Lavoie-Cardinal}]{bilodeau2022microscopy}
\bibinfo{author}{Bilodeau, A.}, \bibinfo{author}{Delmas, C.V.},
  \bibinfo{author}{Parent, M.}, \bibinfo{author}{De~Koninck, P.},
  \bibinfo{author}{Durand, A.}, \bibinfo{author}{Lavoie-Cardinal, F.},
  \bibinfo{year}{2022}.
\newblock \bibinfo{title}{Microscopy analysis neural network to solve
  detection, enumeration and segmentation from image-level annotations}.
\newblock \bibinfo{journal}{Nature Machine Intelligence} \bibinfo{volume}{4},
  \bibinfo{pages}{455--466}.
\bibitem[{Chen et~al.(2016)Chen, Qi, Yu and Heng}]{chen2016dcan}
\bibinfo{author}{Chen, H.}, \bibinfo{author}{Qi, X.}, \bibinfo{author}{Yu, L.},
  \bibinfo{author}{Heng, P.A.}, \bibinfo{year}{2016}.
\newblock \bibinfo{title}{Dcan: deep contour-aware networks for accurate gland
  segmentation}, in: \bibinfo{booktitle}{Proceedings of the IEEE conference on
  Computer Vision and Pattern Recognition}, pp. \bibinfo{pages}{2487--2496}.
\bibitem[{Chen et~al.(2021)Chen, Yuan, Zeng and Wang}]{chen2021semi}
\bibinfo{author}{Chen, X.}, \bibinfo{author}{Yuan, Y.}, \bibinfo{author}{Zeng,
  G.}, \bibinfo{author}{Wang, J.}, \bibinfo{year}{2021}.
\newblock \bibinfo{title}{Semi-supervised semantic segmentation with cross
  pseudo supervision}, in: \bibinfo{booktitle}{Proceedings of the IEEE/CVF
  conference on computer vision and pattern recognition}, pp.
  \bibinfo{pages}{2613--2622}.
\bibitem[{Chen et~al.(2025)Chen, Hou, Lin, Wang, Bie, Wang, Zhou, Chan and
  Chen}]{chen2025segment}
\bibinfo{author}{Chen, Z.}, \bibinfo{author}{Hou, J.}, \bibinfo{author}{Lin,
  L.}, \bibinfo{author}{Wang, Y.}, \bibinfo{author}{Bie, Y.},
  \bibinfo{author}{Wang, X.}, \bibinfo{author}{Zhou, Y.},
  \bibinfo{author}{Chan, R.C.K.}, \bibinfo{author}{Chen, H.},
  \bibinfo{year}{2025}.
\newblock \bibinfo{title}{Segment anything in pathology images with natural
  language}.
\newblock \bibinfo{journal}{arXiv preprint arXiv:2506.20988} .
\bibitem[{Di et~al.(2022)Di, Zou, Feng, Zhou, Ji, Dai and
  Gao}]{di2022generating}
\bibinfo{author}{Di, D.}, \bibinfo{author}{Zou, C.}, \bibinfo{author}{Feng,
  Y.}, \bibinfo{author}{Zhou, H.}, \bibinfo{author}{Ji, R.},
  \bibinfo{author}{Dai, Q.}, \bibinfo{author}{Gao, Y.}, \bibinfo{year}{2022}.
\newblock \bibinfo{title}{Generating hypergraph-based high-order
  representations of whole-slide histopathological images for survival
  prediction}.
\newblock \bibinfo{journal}{IEEE Transactions on Pattern Analysis and Machine
  Intelligence} \bibinfo{volume}{45}, \bibinfo{pages}{5800--5815}.
\bibitem[{Doan et~al.(2022)Doan, Song, Vuong, Kim and Kwak}]{doan2022sonnet}
\bibinfo{author}{Doan, T.N.}, \bibinfo{author}{Song, B.},
  \bibinfo{author}{Vuong, T.T.}, \bibinfo{author}{Kim, K.},
  \bibinfo{author}{Kwak, J.T.}, \bibinfo{year}{2022}.
\newblock \bibinfo{title}{Sonnet: A self-guided ordinal regression neural
  network for segmentation and classification of nuclei in large-scale
  multi-tissue histology images}.
\newblock \bibinfo{journal}{IEEE Journal of Biomedical and Health Informatics}
  \bibinfo{volume}{26}, \bibinfo{pages}{3218--3228}.
\bibitem[{Englbrecht et~al.(2021)Englbrecht, Ruider and
  Bausch}]{englbrecht2021automatic}
\bibinfo{author}{Englbrecht, F.}, \bibinfo{author}{Ruider, I.E.},
  \bibinfo{author}{Bausch, A.R.}, \bibinfo{year}{2021}.
\newblock \bibinfo{title}{Automatic image annotation for fluorescent cell
  nuclei segmentation}.
\newblock \bibinfo{journal}{PloS one} \bibinfo{volume}{16},
  \bibinfo{pages}{e0250093}.
\bibitem[{French et~al.(2020)French, Laine, Aila, Mackiewicz and
  Finlayson}]{french2020semi}
\bibinfo{author}{French, G.}, \bibinfo{author}{Laine, S.},
  \bibinfo{author}{Aila, T.}, \bibinfo{author}{Mackiewicz, M.},
  \bibinfo{author}{Finlayson, G.}, \bibinfo{year}{2020}.
\newblock \bibinfo{title}{Semi-supervised semantic segmentation needs strong,
  varied perturbations}, in: \bibinfo{booktitle}{British Machine Vision
  Conference}.
\bibitem[{Gamper et~al.(2019)Gamper, Alemi~Koohbanani, Benet, Khuram and
  Rajpoot}]{gamper2019pannuke}
\bibinfo{author}{Gamper, J.}, \bibinfo{author}{Alemi~Koohbanani, N.},
  \bibinfo{author}{Benet, K.}, \bibinfo{author}{Khuram, A.},
  \bibinfo{author}{Rajpoot, N.}, \bibinfo{year}{2019}.
\newblock \bibinfo{title}{Pannuke: an open pan-cancer histology dataset for
  nuclei instance segmentation and classification}, in:
  \bibinfo{booktitle}{Digital Pathology: 15th European Congress, ECDP 2019,
  Warwick, UK, April 10--13, 2019, Proceedings 15},
  \bibinfo{organization}{Springer}. pp. \bibinfo{pages}{11--19}.
\bibitem[{Graham et~al.(2019a)Graham, Chen, Gamper, Dou, Heng, Snead, Tsang and
  Rajpoot}]{graham2019mild}
\bibinfo{author}{Graham, S.}, \bibinfo{author}{Chen, H.},
  \bibinfo{author}{Gamper, J.}, \bibinfo{author}{Dou, Q.},
  \bibinfo{author}{Heng, P.A.}, \bibinfo{author}{Snead, D.},
  \bibinfo{author}{Tsang, Y.W.}, \bibinfo{author}{Rajpoot, N.},
  \bibinfo{year}{2019}a.
\newblock \bibinfo{title}{Mild-net: Minimal information loss dilated network
  for gland instance segmentation in colon histology images}.
\newblock \bibinfo{journal}{Medical image analysis} \bibinfo{volume}{52},
  \bibinfo{pages}{199--211}.
\bibitem[{Graham et~al.(2021)Graham, Jahanifar, Azam, Nimir, Tsang, Dodd, Hero,
  Sahota, Tank, Benes et~al.}]{graham2021lizard}
\bibinfo{author}{Graham, S.}, \bibinfo{author}{Jahanifar, M.},
  \bibinfo{author}{Azam, A.}, \bibinfo{author}{Nimir, M.},
  \bibinfo{author}{Tsang, Y.W.}, \bibinfo{author}{Dodd, K.},
  \bibinfo{author}{Hero, E.}, \bibinfo{author}{Sahota, H.},
  \bibinfo{author}{Tank, A.}, \bibinfo{author}{Benes, K.}, et~al.,
  \bibinfo{year}{2021}.
\newblock \bibinfo{title}{Lizard: A large-scale dataset for colonic nuclear
  instance segmentation and classification}, in:
  \bibinfo{booktitle}{Proceedings of the IEEE/CVF international conference on
  computer vision}, pp. \bibinfo{pages}{684--693}.
\bibitem[{Graham et~al.(2019b)Graham, Vu, Raza, Azam, Tsang, Kwak and
  Rajpoot}]{graham2019hover}
\bibinfo{author}{Graham, S.}, \bibinfo{author}{Vu, Q.D.},
  \bibinfo{author}{Raza, S.E.A.}, \bibinfo{author}{Azam, A.},
  \bibinfo{author}{Tsang, Y.W.}, \bibinfo{author}{Kwak, J.T.},
  \bibinfo{author}{Rajpoot, N.}, \bibinfo{year}{2019}b.
\newblock \bibinfo{title}{Hover-net: Simultaneous segmentation and
  classification of nuclei in multi-tissue histology images}.
\newblock \bibinfo{journal}{Medical image analysis} \bibinfo{volume}{58},
  \bibinfo{pages}{101563}.
\bibitem[{Guo et~al.(2023)Guo, Xie, Pagnucco and Song}]{guo2023sac}
\bibinfo{author}{Guo, R.}, \bibinfo{author}{Xie, K.},
  \bibinfo{author}{Pagnucco, M.}, \bibinfo{author}{Song, Y.},
  \bibinfo{year}{2023}.
\newblock \bibinfo{title}{Sac-net: Learning with weak and noisy labels in
  histopathology image segmentation}.
\newblock \bibinfo{journal}{Medical Image Analysis} \bibinfo{volume}{86},
  \bibinfo{pages}{102790}.
\bibitem[{Jin et~al.(2024)Jin, Cui, Sun, Song, Zheng, Cao, Wei and
  Su}]{jin2024inter}
\bibinfo{author}{Jin, Q.}, \bibinfo{author}{Cui, H.}, \bibinfo{author}{Sun,
  C.}, \bibinfo{author}{Song, Y.}, \bibinfo{author}{Zheng, J.},
  \bibinfo{author}{Cao, L.}, \bibinfo{author}{Wei, L.}, \bibinfo{author}{Su,
  R.}, \bibinfo{year}{2024}.
\newblock \bibinfo{title}{Inter-and intra-uncertainty based feature aggregation
  model for semi-supervised histopathology image segmentation}.
\newblock \bibinfo{journal}{Expert Systems with Applications}
  \bibinfo{volume}{238}, \bibinfo{pages}{122093}.
\bibitem[{Jin et~al.(2022)Jin, Cui, Sun, Zheng, Wei, Fang, Meng and
  Su}]{jin2022semi}
\bibinfo{author}{Jin, Q.}, \bibinfo{author}{Cui, H.}, \bibinfo{author}{Sun,
  C.}, \bibinfo{author}{Zheng, J.}, \bibinfo{author}{Wei, L.},
  \bibinfo{author}{Fang, Z.}, \bibinfo{author}{Meng, Z.}, \bibinfo{author}{Su,
  R.}, \bibinfo{year}{2022}.
\newblock \bibinfo{title}{Semi-supervised histological image segmentation via
  hierarchical consistency enforcement}, in: \bibinfo{booktitle}{International
  Conference on Medical Image Computing and Computer-Assisted Intervention},
  \bibinfo{organization}{Springer}. pp. \bibinfo{pages}{3--13}.
\bibitem[{Jung et~al.(2010)Jung, Kim, Chae and Oh}]{jung2010unsupervised}
\bibinfo{author}{Jung, C.}, \bibinfo{author}{Kim, C.}, \bibinfo{author}{Chae,
  S.W.}, \bibinfo{author}{Oh, S.}, \bibinfo{year}{2010}.
\newblock \bibinfo{title}{Unsupervised segmentation of overlapped nuclei using
  bayesian classification}.
\newblock \bibinfo{journal}{IEEE Transactions on Biomedical Engineering}
  \bibinfo{volume}{57}, \bibinfo{pages}{2825--2832}.
\bibitem[{Kuhn(1955)}]{kuhn1955hungarian}
\bibinfo{author}{Kuhn, H.W.}, \bibinfo{year}{1955}.
\newblock \bibinfo{title}{The hungarian method for the assignment problem}.
\newblock \bibinfo{journal}{Naval research logistics quarterly}
  \bibinfo{volume}{2}, \bibinfo{pages}{83--97}.
\bibitem[{Kumar et~al.(2019)Kumar, Verma, Anand, Zhou, Onder, Tsougenis, Chen,
  Heng, Li, Hu et~al.}]{kumar2019multi}
\bibinfo{author}{Kumar, N.}, \bibinfo{author}{Verma, R.},
  \bibinfo{author}{Anand, D.}, \bibinfo{author}{Zhou, Y.},
  \bibinfo{author}{Onder, O.F.}, \bibinfo{author}{Tsougenis, E.},
  \bibinfo{author}{Chen, H.}, \bibinfo{author}{Heng, P.A.},
  \bibinfo{author}{Li, J.}, \bibinfo{author}{Hu, Z.}, et~al.,
  \bibinfo{year}{2019}.
\newblock \bibinfo{title}{A multi-organ nucleus segmentation challenge}.
\newblock \bibinfo{journal}{IEEE transactions on medical imaging}
  \bibinfo{volume}{39}, \bibinfo{pages}{1380--1391}.
\bibitem[{Kumar et~al.(2017)Kumar, Verma, Sharma, Bhargava, Vahadane and
  Sethi}]{kumar2017dataset}
\bibinfo{author}{Kumar, N.}, \bibinfo{author}{Verma, R.},
  \bibinfo{author}{Sharma, S.}, \bibinfo{author}{Bhargava, S.},
  \bibinfo{author}{Vahadane, A.}, \bibinfo{author}{Sethi, A.},
  \bibinfo{year}{2017}.
\newblock \bibinfo{title}{A dataset and a technique for generalized nuclear
  segmentation for computational pathology}.
\newblock \bibinfo{journal}{IEEE transactions on medical imaging}
  \bibinfo{volume}{36}, \bibinfo{pages}{1550--1560}.
\bibitem[{Laine and Aila(2022)}]{laine2022temporal}
\bibinfo{author}{Laine, S.}, \bibinfo{author}{Aila, T.}, \bibinfo{year}{2022}.
\newblock \bibinfo{title}{Temporal ensembling for semi-supervised learning},
  in: \bibinfo{booktitle}{International Conference on Learning
  Representations}.
\bibitem[{Lee et~al.(2013)}]{lee2013pseudo}
\bibinfo{author}{Lee, D.H.}, et~al., \bibinfo{year}{2013}.
\newblock \bibinfo{title}{Pseudo-label: The simple and efficient
  semi-supervised learning method for deep neural networks}, in:
  \bibinfo{booktitle}{Workshop on challenges in representation learning, ICML},
  \bibinfo{organization}{Atlanta}. p. \bibinfo{pages}{896}.
\bibitem[{Li et~al.(2020)Li, Yu, Chen, Fu, Xing and
  Heng}]{li2020transformation}
\bibinfo{author}{Li, X.}, \bibinfo{author}{Yu, L.}, \bibinfo{author}{Chen, H.},
  \bibinfo{author}{Fu, C.W.}, \bibinfo{author}{Xing, L.},
  \bibinfo{author}{Heng, P.A.}, \bibinfo{year}{2020}.
\newblock \bibinfo{title}{Transformation-consistent self-ensembling model for
  semisupervised medical image segmentation}.
\newblock \bibinfo{journal}{IEEE transactions on neural networks and learning
  systems} \bibinfo{volume}{32}, \bibinfo{pages}{523--534}.
\bibitem[{Litjens et~al.(2017)Litjens, Kooi, Bejnordi, Setio, Ciompi,
  Ghafoorian, Van Der~Laak, Van~Ginneken and S{\'a}nchez}]{litjens2017survey}
\bibinfo{author}{Litjens, G.}, \bibinfo{author}{Kooi, T.},
  \bibinfo{author}{Bejnordi, B.E.}, \bibinfo{author}{Setio, A.A.A.},
  \bibinfo{author}{Ciompi, F.}, \bibinfo{author}{Ghafoorian, M.},
  \bibinfo{author}{Van Der~Laak, J.A.}, \bibinfo{author}{Van~Ginneken, B.},
  \bibinfo{author}{S{\'a}nchez, C.I.}, \bibinfo{year}{2017}.
\newblock \bibinfo{title}{A survey on deep learning in medical image analysis}.
\newblock \bibinfo{journal}{Medical image analysis} \bibinfo{volume}{42},
  \bibinfo{pages}{60--88}.
\bibitem[{Liu et~al.(2022)Liu, Tian, Chen, Liu, Belagiannis and
  Carneiro}]{liu2022perturbed}
\bibinfo{author}{Liu, Y.}, \bibinfo{author}{Tian, Y.}, \bibinfo{author}{Chen,
  Y.}, \bibinfo{author}{Liu, F.}, \bibinfo{author}{Belagiannis, V.},
  \bibinfo{author}{Carneiro, G.}, \bibinfo{year}{2022}.
\newblock \bibinfo{title}{Perturbed and strict mean teachers for
  semi-supervised semantic segmentation}, in: \bibinfo{booktitle}{Proceedings
  of the IEEE/CVF conference on computer vision and pattern recognition}, pp.
  \bibinfo{pages}{4258--4267}.
\bibitem[{Lou et~al.(2022)Lou, Li, Li, Han and Wan}]{lou2022pixel}
\bibinfo{author}{Lou, W.}, \bibinfo{author}{Li, H.}, \bibinfo{author}{Li, G.},
  \bibinfo{author}{Han, X.}, \bibinfo{author}{Wan, X.}, \bibinfo{year}{2022}.
\newblock \bibinfo{title}{Which pixel to annotate: a label-efficient nuclei
  segmentation framework}.
\newblock \bibinfo{journal}{IEEE Transactions on Medical Imaging}
  \bibinfo{volume}{42}, \bibinfo{pages}{947--958}.
\bibitem[{Lou et~al.(2024)Lou, Wan, Li, Lou, Li, Gao and Li}]{lou2024structure}
\bibinfo{author}{Lou, W.}, \bibinfo{author}{Wan, X.}, \bibinfo{author}{Li, G.},
  \bibinfo{author}{Lou, X.}, \bibinfo{author}{Li, C.}, \bibinfo{author}{Gao,
  F.}, \bibinfo{author}{Li, H.}, \bibinfo{year}{2024}.
\newblock \bibinfo{title}{Structure embedded nucleus classification for
  histopathology images}.
\newblock \bibinfo{journal}{IEEE Transactions on Medical Imaging} .
\bibitem[{Ma et~al.(2024)Ma, He, Li, Han, You and Wang}]{ma2024segment}
\bibinfo{author}{Ma, J.}, \bibinfo{author}{He, Y.}, \bibinfo{author}{Li, F.},
  \bibinfo{author}{Han, L.}, \bibinfo{author}{You, C.}, \bibinfo{author}{Wang,
  B.}, \bibinfo{year}{2024}.
\newblock \bibinfo{title}{Segment anything in medical images}.
\newblock \bibinfo{journal}{Nature Communications} \bibinfo{volume}{15},
  \bibinfo{pages}{654}.
\bibitem[{Mittal et~al.(2019)Mittal, Tatarchenko and Brox}]{mittal2019semi}
\bibinfo{author}{Mittal, S.}, \bibinfo{author}{Tatarchenko, M.},
  \bibinfo{author}{Brox, T.}, \bibinfo{year}{2019}.
\newblock \bibinfo{title}{Semi-supervised semantic segmentation with high-and
  low-level consistency}.
\newblock \bibinfo{journal}{IEEE transactions on pattern analysis and machine
  intelligence} \bibinfo{volume}{43}, \bibinfo{pages}{1369--1379}.
\bibitem[{Naik et~al.(2008)Naik, Doyle, Agner, Madabhushi, Feldman and
  Tomaszewski}]{naik2008automated}
\bibinfo{author}{Naik, S.}, \bibinfo{author}{Doyle, S.},
  \bibinfo{author}{Agner, S.}, \bibinfo{author}{Madabhushi, A.},
  \bibinfo{author}{Feldman, M.}, \bibinfo{author}{Tomaszewski, J.},
  \bibinfo{year}{2008}.
\newblock \bibinfo{title}{Automated gland and nuclei segmentation for grading
  of prostate and breast cancer histopathology}, in: \bibinfo{booktitle}{2008
  5th IEEE International Symposium on Biomedical Imaging: From Nano to Macro},
  \bibinfo{organization}{IEEE}. pp. \bibinfo{pages}{284--287}.
\bibitem[{Nunes et~al.(2025)Nunes, Montezuma, Oliveira, Pereira and
  Cardoso}]{nunes2025survey}
\bibinfo{author}{Nunes, J.D.}, \bibinfo{author}{Montezuma, D.},
  \bibinfo{author}{Oliveira, D.}, \bibinfo{author}{Pereira, T.},
  \bibinfo{author}{Cardoso, J.S.}, \bibinfo{year}{2025}.
\newblock \bibinfo{title}{A survey on cell nuclei instance segmentation and
  classification: Leveraging context and attention}.
\newblock \bibinfo{journal}{Medical Image Analysis} \bibinfo{volume}{99},
  \bibinfo{pages}{103360}.
\bibitem[{Ouali et~al.(2020)Ouali, Hudelot and Tami}]{ouali2020semi}
\bibinfo{author}{Ouali, Y.}, \bibinfo{author}{Hudelot, C.},
  \bibinfo{author}{Tami, M.}, \bibinfo{year}{2020}.
\newblock \bibinfo{title}{Semi-supervised semantic segmentation with
  cross-consistency training}, in: \bibinfo{booktitle}{Proceedings of the
  IEEE/CVF conference on computer vision and pattern recognition}, pp.
  \bibinfo{pages}{12674--12684}.
\bibitem[{Pan et~al.(2023)Pan, Cheng, Hou, Lan, Lu, Li, Feng, Wang, Liang, Liu
  et~al.}]{pan2023smile}
\bibinfo{author}{Pan, X.}, \bibinfo{author}{Cheng, J.}, \bibinfo{author}{Hou,
  F.}, \bibinfo{author}{Lan, R.}, \bibinfo{author}{Lu, C.},
  \bibinfo{author}{Li, L.}, \bibinfo{author}{Feng, Z.}, \bibinfo{author}{Wang,
  H.}, \bibinfo{author}{Liang, C.}, \bibinfo{author}{Liu, Z.}, et~al.,
  \bibinfo{year}{2023}.
\newblock \bibinfo{title}{Smile: Cost-sensitive multi-task learning for nuclear
  segmentation and classification with imbalanced annotations}.
\newblock \bibinfo{journal}{Medical Image Analysis} \bibinfo{volume}{88},
  \bibinfo{pages}{102867}.
\bibitem[{Qu et~al.(2019)Qu, Yan, Riedlinger, De and Metaxas}]{qu2019improving}
\bibinfo{author}{Qu, H.}, \bibinfo{author}{Yan, Z.},
  \bibinfo{author}{Riedlinger, G.M.}, \bibinfo{author}{De, S.},
  \bibinfo{author}{Metaxas, D.N.}, \bibinfo{year}{2019}.
\newblock \bibinfo{title}{Improving nuclei/gland instance segmentation in
  histopathology images by full resolution neural network and spatial
  constrained loss}, in: \bibinfo{booktitle}{Medical Image Computing and
  Computer Assisted Intervention--MICCAI 2019: 22nd International Conference,
  Shenzhen, China, October 13--17, 2019, Proceedings, Part I 22},
  \bibinfo{organization}{Springer}. pp. \bibinfo{pages}{378--386}.
\bibitem[{Sharma et~al.(2015)Sharma, Zerbe, Heim, Wienert, Behrens, Hellwich
  and Hufnagl}]{sharma2015multi}
\bibinfo{author}{Sharma, H.}, \bibinfo{author}{Zerbe, N.},
  \bibinfo{author}{Heim, D.}, \bibinfo{author}{Wienert, S.},
  \bibinfo{author}{Behrens, H.M.}, \bibinfo{author}{Hellwich, O.},
  \bibinfo{author}{Hufnagl, P.}, \bibinfo{year}{2015}.
\newblock \bibinfo{title}{A multi-resolution approach for combining visual
  information using nuclei segmentation and classification in histopathological
  images}, in: \bibinfo{booktitle}{International Conference on Computer Vision
  Theory and Applications}, \bibinfo{organization}{SCITEPRESS}. pp.
  \bibinfo{pages}{37--46}.
\bibitem[{Shen et~al.(2023)Shen, Cao, Yang, Liu, Yang and Zaiane}]{shen2023co}
\bibinfo{author}{Shen, Z.}, \bibinfo{author}{Cao, P.}, \bibinfo{author}{Yang,
  H.}, \bibinfo{author}{Liu, X.}, \bibinfo{author}{Yang, J.},
  \bibinfo{author}{Zaiane, O.R.}, \bibinfo{year}{2023}.
\newblock \bibinfo{title}{Co-training with high-confidence pseudo labels for
  semi-supervised medical image segmentation}, in:
  \bibinfo{booktitle}{Proceedings of the Thirty-Second International Joint
  Conference on Artificial Intelligence}, pp. \bibinfo{pages}{4199--4207}.
\bibitem[{Sohn et~al.(2020)Sohn, Berthelot, Carlini, Zhang, Zhang, Raffel,
  Cubuk, Kurakin and Li}]{sohn2020fixmatch}
\bibinfo{author}{Sohn, K.}, \bibinfo{author}{Berthelot, D.},
  \bibinfo{author}{Carlini, N.}, \bibinfo{author}{Zhang, Z.},
  \bibinfo{author}{Zhang, H.}, \bibinfo{author}{Raffel, C.A.},
  \bibinfo{author}{Cubuk, E.D.}, \bibinfo{author}{Kurakin, A.},
  \bibinfo{author}{Li, C.L.}, \bibinfo{year}{2020}.
\newblock \bibinfo{title}{Fixmatch: Simplifying semi-supervised learning with
  consistency and confidence}.
\newblock \bibinfo{journal}{Advances in neural information processing systems}
  \bibinfo{volume}{33}, \bibinfo{pages}{596--608}.
\bibitem[{Taha and Hanbury(2015)}]{taha2015metrics}
\bibinfo{author}{Taha, A.A.}, \bibinfo{author}{Hanbury, A.},
  \bibinfo{year}{2015}.
\newblock \bibinfo{title}{Metrics for evaluating 3d medical image segmentation:
  analysis, selection, and tool}.
\newblock \bibinfo{journal}{BMC medical imaging} \bibinfo{volume}{15},
  \bibinfo{pages}{1--28}.
\bibitem[{Tarvainen and Valpola(2017)}]{tarvainen2017mean}
\bibinfo{author}{Tarvainen, A.}, \bibinfo{author}{Valpola, H.},
  \bibinfo{year}{2017}.
\newblock \bibinfo{title}{Mean teachers are better role models: Weight-averaged
  consistency targets improve semi-supervised deep learning results}.
\newblock \bibinfo{journal}{Advances in neural information processing systems}
  \bibinfo{volume}{30}.
\bibitem[{Verma et~al.(2021)Verma, Kumar, Patil, Kurian, Rane, Graham, Vu,
  Zwager, Raza, Rajpoot et~al.}]{verma2021monusac2020}
\bibinfo{author}{Verma, R.}, \bibinfo{author}{Kumar, N.},
  \bibinfo{author}{Patil, A.}, \bibinfo{author}{Kurian, N.C.},
  \bibinfo{author}{Rane, S.}, \bibinfo{author}{Graham, S.},
  \bibinfo{author}{Vu, Q.D.}, \bibinfo{author}{Zwager, M.},
  \bibinfo{author}{Raza, S.E.A.}, \bibinfo{author}{Rajpoot, N.}, et~al.,
  \bibinfo{year}{2021}.
\newblock \bibinfo{title}{Monusac2020: A multi-organ nuclei segmentation and
  classification challenge}.
\newblock \bibinfo{journal}{IEEE Transactions on Medical Imaging}
  \bibinfo{volume}{40}, \bibinfo{pages}{3413--3423}.
\bibitem[{Veta et~al.(2013)Veta, Van~Diest, Kornegoor, Huisman, Viergever and
  Pluim}]{veta2013automatic}
\bibinfo{author}{Veta, M.}, \bibinfo{author}{Van~Diest, P.J.},
  \bibinfo{author}{Kornegoor, R.}, \bibinfo{author}{Huisman, A.},
  \bibinfo{author}{Viergever, M.A.}, \bibinfo{author}{Pluim, J.P.},
  \bibinfo{year}{2013}.
\newblock \bibinfo{title}{Automatic nuclei segmentation in h\&e stained breast
  cancer histopathology images}.
\newblock \bibinfo{journal}{PloS one} \bibinfo{volume}{8},
  \bibinfo{pages}{e70221}.
\bibitem[{Vincent and Soille(1991)}]{vincent1991watersheds}
\bibinfo{author}{Vincent, L.}, \bibinfo{author}{Soille, P.},
  \bibinfo{year}{1991}.
\newblock \bibinfo{title}{Watersheds in digital spaces: an efficient algorithm
  based on immersion simulations}.
\newblock \bibinfo{journal}{IEEE Transactions on Pattern Analysis \& Machine
  Intelligence} \bibinfo{volume}{13}, \bibinfo{pages}{583--598}.
\bibitem[{Wu et~al.(2022)Wu, Wang, Song, Yang and Qin}]{wu2022cross}
\bibinfo{author}{Wu, H.}, \bibinfo{author}{Wang, Z.}, \bibinfo{author}{Song,
  Y.}, \bibinfo{author}{Yang, L.}, \bibinfo{author}{Qin, J.},
  \bibinfo{year}{2022}.
\newblock \bibinfo{title}{Cross-patch dense contrastive learning for
  semi-supervised segmentation of cellular nuclei in histopathologic images},
  in: \bibinfo{booktitle}{Proceedings of the IEEE/CVF Conference on Computer
  Vision and Pattern Recognition}, pp. \bibinfo{pages}{11666--11675}.
\bibitem[{Wu et~al.(2025)Wu, Wang, Hong, Ji, Fu, Xu, Xu and
  Jin}]{wu2025medical}
\bibinfo{author}{Wu, J.}, \bibinfo{author}{Wang, Z.}, \bibinfo{author}{Hong,
  M.}, \bibinfo{author}{Ji, W.}, \bibinfo{author}{Fu, H.}, \bibinfo{author}{Xu,
  Y.}, \bibinfo{author}{Xu, M.}, \bibinfo{author}{Jin, Y.},
  \bibinfo{year}{2025}.
\newblock \bibinfo{title}{Medical sam adapter: Adapting segment anything model
  for medical image segmentation}.
\newblock \bibinfo{journal}{Medical image analysis} \bibinfo{volume}{102},
  \bibinfo{pages}{103547}.
\bibitem[{Xie et~al.(2021)Xie, Zhang, Liao, Verjans, Shen and
  Xia}]{xie2021intra}
\bibinfo{author}{Xie, Y.}, \bibinfo{author}{Zhang, J.}, \bibinfo{author}{Liao,
  Z.}, \bibinfo{author}{Verjans, J.}, \bibinfo{author}{Shen, C.},
  \bibinfo{author}{Xia, Y.}, \bibinfo{year}{2021}.
\newblock \bibinfo{title}{Intra-and inter-pair consistency for semi-supervised
  gland segmentation}.
\newblock \bibinfo{journal}{IEEE Transactions on Image Processing}
  \bibinfo{volume}{31}, \bibinfo{pages}{894--905}.
\bibitem[{Yang et~al.(2006)Yang, Li and Zhou}]{yang2006nuclei}
\bibinfo{author}{Yang, X.}, \bibinfo{author}{Li, H.}, \bibinfo{author}{Zhou,
  X.}, \bibinfo{year}{2006}.
\newblock \bibinfo{title}{Nuclei segmentation using marker-controlled
  watershed, tracking using mean-shift, and kalman filter in time-lapse
  microscopy}.
\newblock \bibinfo{journal}{IEEE Transactions on Circuits and Systems I:
  Regular Papers} \bibinfo{volume}{53}, \bibinfo{pages}{2405--2414}.
\bibitem[{Yu et~al.(2019)Yu, Wang, Li, Fu and Heng}]{yu2019uncertainty}
\bibinfo{author}{Yu, L.}, \bibinfo{author}{Wang, S.}, \bibinfo{author}{Li, X.},
  \bibinfo{author}{Fu, C.W.}, \bibinfo{author}{Heng, P.A.},
  \bibinfo{year}{2019}.
\newblock \bibinfo{title}{Uncertainty-aware self-ensembling model for
  semi-supervised 3d left atrium segmentation}, in: \bibinfo{booktitle}{Medical
  image computing and computer assisted intervention--MICCAI 2019: 22nd
  international conference, Shenzhen, China, October 13--17, 2019, proceedings,
  part II 22}, \bibinfo{organization}{Springer}. pp. \bibinfo{pages}{605--613}.
\bibitem[{Zhang and Liu(2023)}]{zhang2023customized}
\bibinfo{author}{Zhang, K.}, \bibinfo{author}{Liu, D.}, \bibinfo{year}{2023}.
\newblock \bibinfo{title}{Customized segment anything model for medical image
  segmentation}.
\newblock \bibinfo{journal}{arXiv preprint arXiv:2304.13785} .
\bibitem[{Zhang et~al.(2024)Zhang, Yuan, Zhou, Wang, Chen and
  Wang}]{zhang2024venet}
\bibinfo{author}{Zhang, S.}, \bibinfo{author}{Yuan, Z.}, \bibinfo{author}{Zhou,
  X.}, \bibinfo{author}{Wang, H.}, \bibinfo{author}{Chen, B.},
  \bibinfo{author}{Wang, Y.}, \bibinfo{year}{2024}.
\newblock \bibinfo{title}{Venet: Variational energy network for gland
  segmentation of pathological images and early gastric cancer diagnosis of
  whole slide images}.
\newblock \bibinfo{journal}{Computer Methods and Programs in Biomedicine}
  \bibinfo{volume}{250}, \bibinfo{pages}{108178}.
\bibitem[{Zhou et~al.(2023)Zhou, Wu, Wang, Wei, Lai, Shou, Fan and
  Xu}]{zhou2023cyclic}
\bibinfo{author}{Zhou, Y.}, \bibinfo{author}{Wu, Y.}, \bibinfo{author}{Wang,
  Z.}, \bibinfo{author}{Wei, B.}, \bibinfo{author}{Lai, M.},
  \bibinfo{author}{Shou, J.}, \bibinfo{author}{Fan, Y.}, \bibinfo{author}{Xu,
  Y.}, \bibinfo{year}{2023}.
\newblock \bibinfo{title}{Cyclic learning: Bridging image-level labels and
  nuclei instance segmentation}.
\newblock \bibinfo{journal}{IEEE Transactions on Medical Imaging}
  \bibinfo{volume}{42}, \bibinfo{pages}{3104--3116}.
\bibitem[{Zhu et~al.(2021)Zhu, Zhang, Wu, Zhang, He, Zhang, Manmatha, Li and
  Smola}]{zhu2021improving}
\bibinfo{author}{Zhu, Y.}, \bibinfo{author}{Zhang, Z.}, \bibinfo{author}{Wu,
  C.}, \bibinfo{author}{Zhang, Z.}, \bibinfo{author}{He, T.},
  \bibinfo{author}{Zhang, H.}, \bibinfo{author}{Manmatha, R.},
  \bibinfo{author}{Li, M.}, \bibinfo{author}{Smola, A.}, \bibinfo{year}{2021}.
\newblock \bibinfo{title}{Improving semantic segmentation via efficient
  self-training}.
\newblock \bibinfo{journal}{IEEE transactions on pattern analysis and machine
  intelligence} \bibinfo{volume}{46}, \bibinfo{pages}{1589--1602}.

\end{thebibliography}
\end{document}